\documentclass{article}
\usepackage{microtype}
\usepackage{graphicx}
\usepackage{subcaption}
\usepackage{booktabs} 

\usepackage{hyperref}

\usepackage{tcolorbox}
\tcbuselibrary{breakable}

\usepackage[preprint]{icml2026}

\usepackage{amsmath}
\usepackage{amssymb}
\usepackage{mathtools}
\usepackage{amsthm}
\usepackage{enumitem}
\usepackage{adjustbox}
\usepackage{multirow}
\usepackage{xurl}

\usepackage[capitalize,noabbrev]{cleveref}

\theoremstyle{plain}

\theoremstyle{definition}

\theoremstyle{remark}

\usepackage[textsize=tiny]{todonotes}

\icmltitlerunning{Self-Verification Dilemma: Experience-Driven Suppression of Overused Checking in LLM Reasoning}

\begin{document}

\twocolumn[
  \icmltitle{Self-Verification Dilemma: Experience-Driven Suppression of \\Overused Checking in LLM Reasoning}



  \icmlsetsymbol{equal}{*}



  \begin{icmlauthorlist}
    \icmlauthor{Quanyu Long}{ntu}
    \icmlauthor{Kai Jie Jiang}{ntu}
    \icmlauthor{Jianda Chen}{ntu}
    \icmlauthor{Xu Guo}{ntu}
    \icmlauthor{Leilei Gan}{zju}
    \icmlauthor{Wenya Wang}{ntu}
  \end{icmlauthorlist}

  \icmlaffiliation{ntu}{Nanyang Technological University}
  \icmlaffiliation{zju}{Zhejiang University}

  \icmlcorrespondingauthor{Quanyu Long}{quanyu001@e.ntu.edu.sg}
  \icmlcorrespondingauthor{Wenya Wang}{wangwy@ntu.edu.sg}
  \icmlkeywords{Machine Learning, ICML}

  \vskip 0.3in
]



\printAffiliationsAndNotice{}  

\begin{abstract}

Large Reasoning Models (LRMs) achieve strong performance by generating long reasoning traces with reflection. Through a large-scale empirical analysis, we find that a substantial fraction of reflective steps consist of self-verification (recheck) that repeatedly confirm intermediate results. These rechecks occur frequently across models and benchmarks, yet the vast majority are confirmatory rather than corrective, rarely identifying errors and altering reasoning outcomes. This reveals a mismatch between how often self-verification is activated and how often it is actually useful. Motivated by this, we propose a novel, experience-driven test-time framework that reduces the overused verification. Our method detects the activation of recheck behavior, consults an offline experience pool of past verification outcomes, and estimates whether a recheck is likely unnecessary via efficient retrieval. When historical experience suggests unnecessary, a suppression signal redirects the model to proceed. Across multiple model and benchmarks, our approach reduces token usage up to 20.3 \% while maintaining the accuracy, and in some datasets even yields accuracy improvements.
\end{abstract}

\section{Introduction}
\label{sec:intro}
\begin{figure}[t]
	\centering
    \setlength{\belowcaptionskip}{-0.5cm}
	\includegraphics[width=0.85\columnwidth]{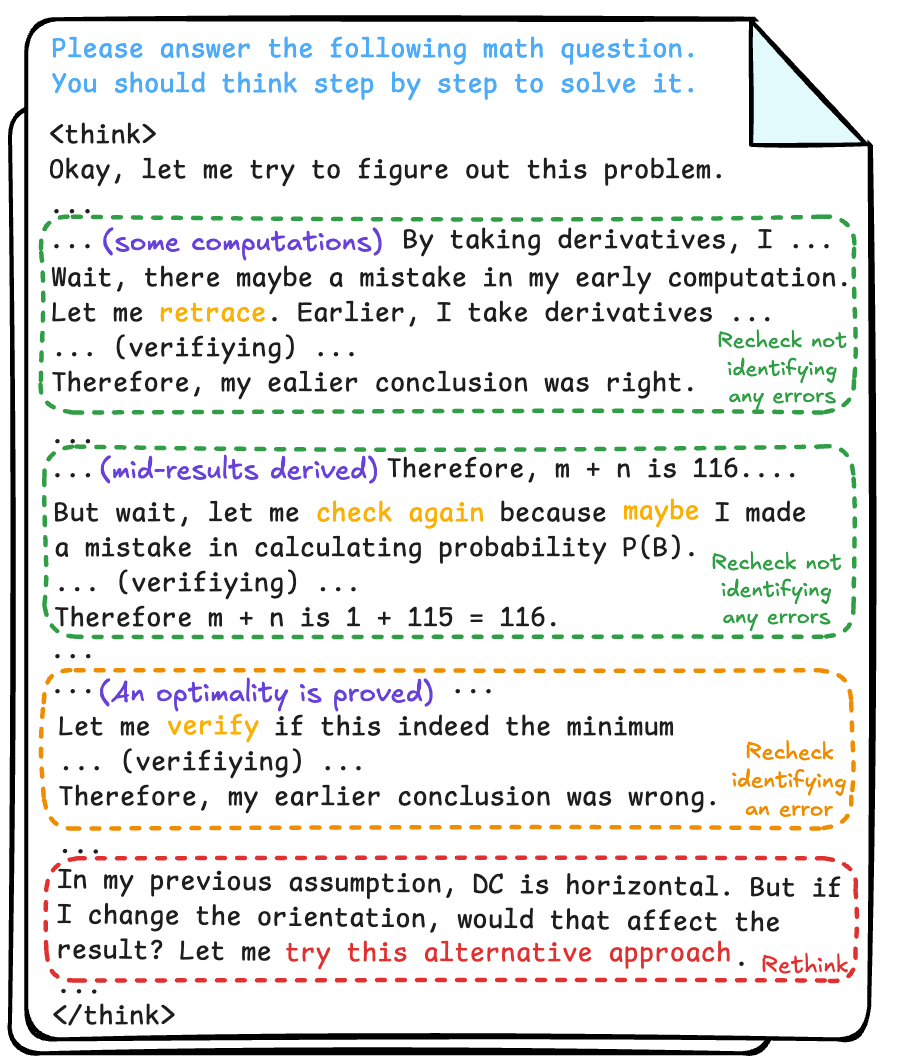}

       \caption{Reflective behaviors commonly observed in step-by-step mathematical reasoning. We illustrate three categories: rethink, where the model revises its strategy and explores an alternative line of reasoning; and recheck, where the model verifies already-derived intermediate results through re-computation, arithmetic checking, or optimality validation. Recheck behaviors can be further divided into confirmatory cases, which only confirm previous results, and corrective cases, which identify errors and alter the reasoning outcome. Our empirical analysis shows that recheck behaviors overwhelmingly fall into the confirmatory category.}

	\label{fig:intro}
\end{figure}


Large language models (LLMs) have recently demonstrated substantial progress on complex reasoning tasks. Large Reasoning Models (LRMs) such as OpenAI o1~\citep{openai2024openaio1card} and DeepSeek-R1 \citep{deepseekai2025deepseekr1incentivizingreasoningcapability} further extend these capabilities through reinforcement learning (RL), producing long reasoning traces that exhibit reflection and self-correction in an increasingly anthropomorphic manner \citep{yang2025understandingahamomentsexternal}. A useful lens is that RL does not introduce reflection from scratch but scales it up: pretrained LLMs only sporadically exhibit reflective behaviors, and these are often shallow and unlikely to improve solutions \citep{liu2025oatzero}, whereas RL markedly increases both the frequency and apparent effectiveness of reflection by optimizing task-level rewards. This makes reflection a key mechanism for improving reliability in LRMs. However, the same mechanism can also lead to inefficient computation, where models continue to deliberate and repeatedly verify, despite already having sufficient evidence, a phenomenon often described as overthinking~\citep{sui2025stopoverthinkingsurveyefficient}. Together, these observations position reflection as a central but double-edged component of LRMs.

While overthinking in reasoning models has been increasingly studied \citep{zhang-etal-2025-adaptthink, fang2025thinkless, jiang2025think, zhang2026othinkr1intrinsicfastslowthinking}, we observe that a large fraction of reflective steps in practice consist of repeated and often trivial rechecks (Figure~\ref{fig:intro}). Through large-scale analysis of reasoning traces, we find that reflective behaviors can be broadly categorized into two types: 1) \textbf{rethink}, where the model revises its reasoning strategy or reformulates the problem, and 2) \textbf{recheck} (i.e., \textbf{self-verification}), where the model verifies the correctness of already-derived intermediate results through re-computation, arithmetic checking, constraint validation, or optimality testing. 
Our analysis further shows that recheck occurs \textbf{frequently} during model reflective reasoning across models and datasets (Section~\ref{sec:second_ana}).
However, most rechecks are \textbf{confirmatory}: they rarely identify errors or change the reasoning outcome (Section~\ref{sec:third_ana}).

These observations are consistent with findings from \citet{kang2025trymattersrevisitingrole}, who shows that verification steps following a candidate final answer rarely change the final result; 
in contrast, we study self-verification as a manifestation of the model’s epistemic uncertainty during the reasoning process itself, rather than as post hoc confirmation of a final answer. 
The prevalence of such redundant rechecking presents a practical challenge: although intended to improve reliability, excessive self-verification consumes substantial reasoning budget and yields diminishing corrective benefit, revealing a mismatch between how often rechecking is performed and how often it is actually useful.

This raises a central question: \textbf{when is self-verification actually worth performing during reasoning?} 
In human problem solving, self-verification is rarely applied uniformly. When encountering a familiar operation, such as taking derivatives of a standard expression or checking a routine constraint, humans often rely on accumulated confidence and proceed without rechecking every step, knowing that similar derivations have rarely failed in the past.
Inspired by this, we argue that effective reasoning requires selective self-verification: verifications should be executed aggressively in contexts where errors are likely and conservatively when prior evidence suggests high confidence. Rather than treating self-verification as an unconditional safeguard, it should be guided by experience, allowing the reasoning process to allocate effort where it is most likely to be beneficial.

Motivated by this perspective, we propose an experience-driven framework for selective self-verification at test-time. 
Our approach equips a reasoning model with access to past self-verification experience, allowing it to estimate whether a newly initiated verification step is likely to be useful. Concretely, during inference, we first detect the activation of self-verification behaviors using a lightweight trained  identifier. Upon detection, we retrieve similar verification episodes from an experience pool constructed offline. By aggregating historical outcomes, specifically, whether similar rechecks on a certain concrete mathematical manipulations (e.g., taking derivatives) previously corrected errors or merely confirmed correctness, then we can estimate the likelihood that continuing the current verification will be beneficial by a voting strategy. When prior experience indicates that rechecking is unlikely to change the reasoning outcome, we suppress the verification by injecting a brief textual signal and allow the model to proceed to subsequent reasoning steps. Importantly, our method does not modify model parameters, does not truncate necessary reasoning and strategy-level rethinking, and intervenes only on redundant rechecks. Empirically, this selective suppression reduces token usage by up to 20.3\% while maintaining comparable accuracy, and in some datasets even yields slight accuracy improvements. 

\section{Related Work}
\label{sec:related}
\vspace{-0.5em}
\paragraph{Self-Evolving LLM with Experience}
Recent work regarding enabling LLMs to self-evolve from experience focuses on mechanisms for accumulating, abstracting, and reusing interaction histories. ExpeL~\citep{zhao2024expel} and Contextual Experience Replay~\citep{liu-etal-2025-contextual} introduce training-free experiential learning by distilling and retrieving past interaction trajectories to improve future performance, while Retrospex~\citep{yufei2024retrospex} further integrates experience via an offline RL critic to guide action selection. AgentEvolver~\citep{AgentEvolver2025} extends this paradigm by combining self-questioning, self-navigation, and self-attribution to enable continual autonomous improvement. Complementary studies analyze the limits of test-time learning from experience, showing measurable but slower gains compared to humans~\citep{wang-etal-2025-far}. Other approaches focus on structuring experience for reuse: HiAgent~\citep{hu-etal-2025-hiagent} hierarchically manages working memory for long-horizon tasks, while Metacognitive Reuse~\citep{anonymous2026metacognitive} compresses recurring reasoning into reusable behaviors. Finally, efficiency-oriented methods optimize experiential reasoning by decomposing or truncating reasoning trajectories, including structured multi-turn decomposition~\citep{anonymous2026done}, self-braking tuning~\citep{zhao2025let}, and early stopping informed by reflection analysis~\citep{kang2025trymattersrevisitingrole}.

\vspace{-1em}
\paragraph{Efficient Reasoning}
Recent work on efficient reasoning increasingly focuses on adaptive compute allocation. A line of methods trains models to decide whether to reason, or to switch between fast and slow modes based on input difficulty, including query-level think/no-think decisions and hybrid reasoning strategies \cite{zhang-etal-2025-adaptthink,fang2025thinkless,jiang2025think,zhang-etal-2025-continue}. This paradigm has been extended to intrinsic fast–slow switching and multi-format reasoning selection \cite{zhang2026othinkr1intrinsicfastslowthinking,wu2025arm}. Finer-grained approaches further adapt computation at the step or process level, for example by leveraging process-level rewards to switch reasoning strategies \cite{wang2025patsprocessleveladaptivethinking}, or by allocating additional computation only to high-value reasoning segments through long–short collaboration \cite{anonymous2026not}. Complementary analyses observe that reflective reasoning frequently reconfirms the initial answer rather than correcting errors, motivating reflection truncation or early-stopping mechanisms \cite{kang2025trymattersrevisitingrole}. In parallel, inference cost has been reduced through adaptive length or budget control and convergence-based stopping criteria \cite{shen-etal-2025-dast,liu-wang-2025-answer,anonymous2026early}, while system-oriented model families provide practical reasoning toggles for high-throughput deployment \cite{bercovich2025llamanemotronefficientreasoningmodels}.
In contrast to prior work that primarily allocates compute by switching reasoning modes or terminating reasoning early based on difficulty, our approach targets redundant self-verification during inference. We selectively suppress low-utility rechecking behaviors at test time, with reduced reasoning length arising naturally as a consequence rather than an explicit optimization objective.

  

\section{Empirical Analysis}
\label{sec:ana}


All analyses are conducted on rollouts generated by four LLMs, Qwen3-8B \cite{yang2025qwen3}, QwQ-32B \cite{qwq32b}, gpt-oss-20b \cite{openai2025gptoss120bgptoss20bmodel}, and DeepSeek-R1-Distill-Qwen-7B \cite{deepseekai2025deepseekr1incentivizingreasoningcapability}. We evaluate models on four mathematical reasoning benchmarks: AIME24 \cite{AIME2024}, AIME25 \cite{AIME2025}, AMC23 \cite{AMC} and MATH500 \cite{hendrycks2021measuring}.

\vspace{-0.5em}
\subsection{How frequently do LLMs exhibit reflective behaviour?}
\label{sec:first_ana}

Recent LLMs generate long Chain-of-Thoughts (CoTs) that frequently include reflective behaviors. These reflections are commonly viewed as a mechanism for improving reasoning reliability, particularly in models trained to deliberate extensively.
To better understand the practical role of reflection, we begin by quantifying how frequently such behaviors occur in reasoning traces.

\begin{figure}[t]
    \centering
    \setlength{\abovecaptionskip}{0.1cm}
    \setlength{\belowcaptionskip}{-0.4cm}
    \includegraphics[width=0.95\linewidth]{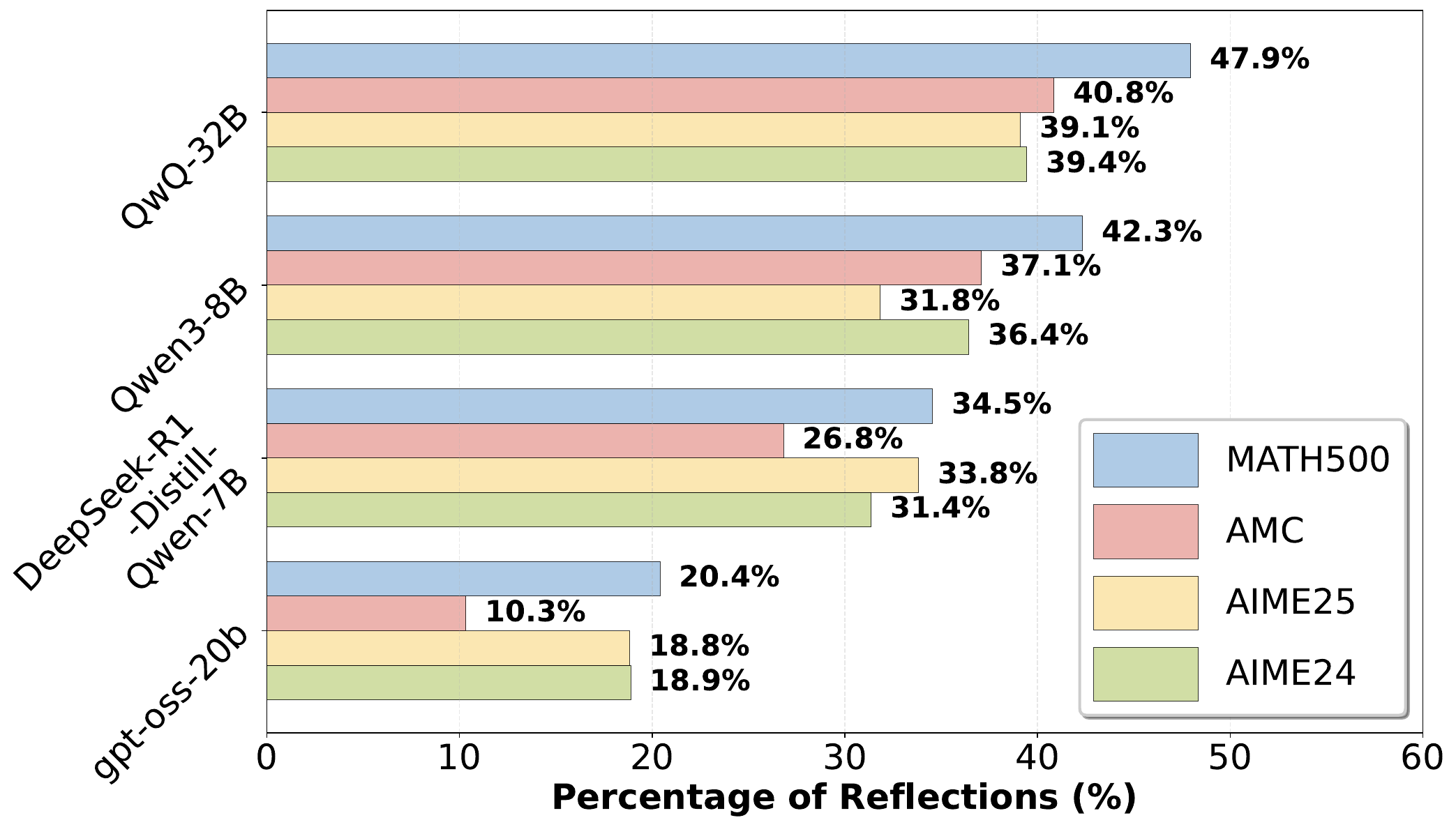}
    \caption{Percentage of steps classified as reflections.}
    \label{fig:percentage_reflection}
\end{figure}


We begin by analyzing reasoning rollouts at the granularity of individual steps. Each rollout is segmented into a sequence of steps using simple formatting cues ``\textbackslash n\textbackslash n''.
Each segmented step is then annotated as either reflective or non-reflective using GPT-5~\citep{singh2025openaigpt5card}. 
The classifier is prompted to determine whether a step constitutes a reflection based on a strict rule: a step is labeled as reflective only if it explicitly refers to the model’s own reasoning process and engages in monitoring, evaluating, or revising that reasoning. Concretely, reflective steps include explicit self-referential or metacognitive language indicating verification, critique, reconsideration, or switching to new reasoning strategies (e.g., ``I should check'', ``this approach might be wrong'', ``let me reconsider''). In contrast, steps that primarily advance the solution, such as performing calculations, applying formulas, or explaining problem-specific logic, are not considered reflective. The exact instructional prompts are provided in Appendix~\ref{sec:appendixE}. Based on these annotations, we compute the proportion of reflective steps within each rollout and present the results in Figure \ref{fig:percentage_reflection}.

\vspace{-0.5em}
\paragraph{Findings.} 
We observe that across all reasoning-oriented models and benchmarks, reflective steps constitute a substantial portion of model-generated rollouts, often approaching or exceeding one third of all steps. These results indicate that reflection is a systematic and frequent component of LLM reasoning.



Interestingly, a common intuition is that reflection should correlate with problem difficulty: harder problems presumably require more exploration, reconsideration, and strategic adjustment, leading to more frequent reflective behavior. However, our results reveal a counterintuitive pattern. For all models, MATH500 exhibits the highest proportion of reflective steps, despite its problems being less difficult than those in AIME.
This finding suggests that \textbf{reflection frequency does not reliably track problem difficulty or the need for deeper reasoning}. Instead, reflection appears to be triggered broadly during reasoning, even in settings where extensive deliberation is unlikely to advance the solution. This motivates a closer examination of what kinds of reflection models actually perform. 


\vspace{-0.5em}
\subsection{What Forms of Reflection do LLMs repeatedly exhibit?}
\label{sec:second_ana}


In practice, reflective steps may serve qualitatively different functional roles. Some reflections support advanced problem solving by revising strategies or restructuring the solution path, while others merely verify previously derived results without introducing new reasoning. Distinguishing between these roles is essential for understanding whether reflection primarily advances reasoning or instead functions as local self-monitoring.

\vspace{-0.5em}
\paragraph{Rethink vs. Recheck.}
We introduce a functional taxonomy that separates reflective steps into two categories: rethink and recheck (Examples are provided in Figure \ref{fig:intro}). \textbf{Rethinks} operate at the level of global reasoning strategy. Rather than evaluating a specific intermediate result, they revise assumptions, introduce alternative solution approaches, abandon earlier plans, or reorganize the structure of the reasoning process. These reflections are inherently exploratory and are often necessary for solving complex problems where the initial approach is flawed or incomplete. \textbf{Rechecks}(i.e., \textbf{self-verification}), in contrast, operate at the level of local correctness. They focus on verifying previously stated intermediate results, computations, or logical steps, with the explicit or implicit goal of assessing whether an earlier conclusion is valid. Rechecks may involve recomputation, constraint checking, case testing, or validation of optimality or minimality. Importantly, rechecks do not introduce new solution strategies; they evaluate existing ones.


To operationalize this taxonomy, for each step identified as reflective in Section \ref{sec:first_ana}, we prompt GPT-5 with surrounding context, including preceding and subsequent steps, enabling it to assess the role that the local reflection plays in the overall reasoning trajectory. Each reflective step is then categorized as either a rethink or a recheck, with ambiguous cases labeled as unable to classify.

\vspace{-0.5em}
\paragraph{Findings.}
Figure \ref{fig:reflection_proportion} presents the distribution of reflection types for Qwen3-8B across four benchmarks. Across all datasets, rechecks constitute a substantial fraction of reflective steps, accounting for approximately 40–58\% of all reflections. Notably, rechecks make up a larger share of reflections on comparatively simpler datasets, such as AMC and MATH500.
This pattern aligns closely with our findings in the previous subsection. There, we show that reflective steps are not more frequent on harder problems, instead, simpler datasets often induce more frequent reflection. The present analysis clarifies that on easier problems, reflection is dominated by local self-verification rather than by meaningful strategy revision. As a result, reflection frequently manifests as repetitive rechecking of intermediate results. This observation motivates a closer examination of whether such rechecks meaningfully contribute to correctness.



\vspace{-0.5em}
\subsection{Do Rechecks Truly Identify Errors?}
\label{sec:third_ana}
If rechecks frequently identify and correct errors, their prevalence would justify self-verification as an essential component of reliable multi-step reasoning. However, if most rechecks simply reaffirm already-correct intermediate results, then a substantial portion of reflective behavior is redundant. We therefore examine whether rechecks in practice meaningfully contribute to error correction.

\begin{figure}[t]
    \centering
    \setlength{\belowcaptionskip}{-0.4cm}
    \includegraphics[width=0.75\linewidth]{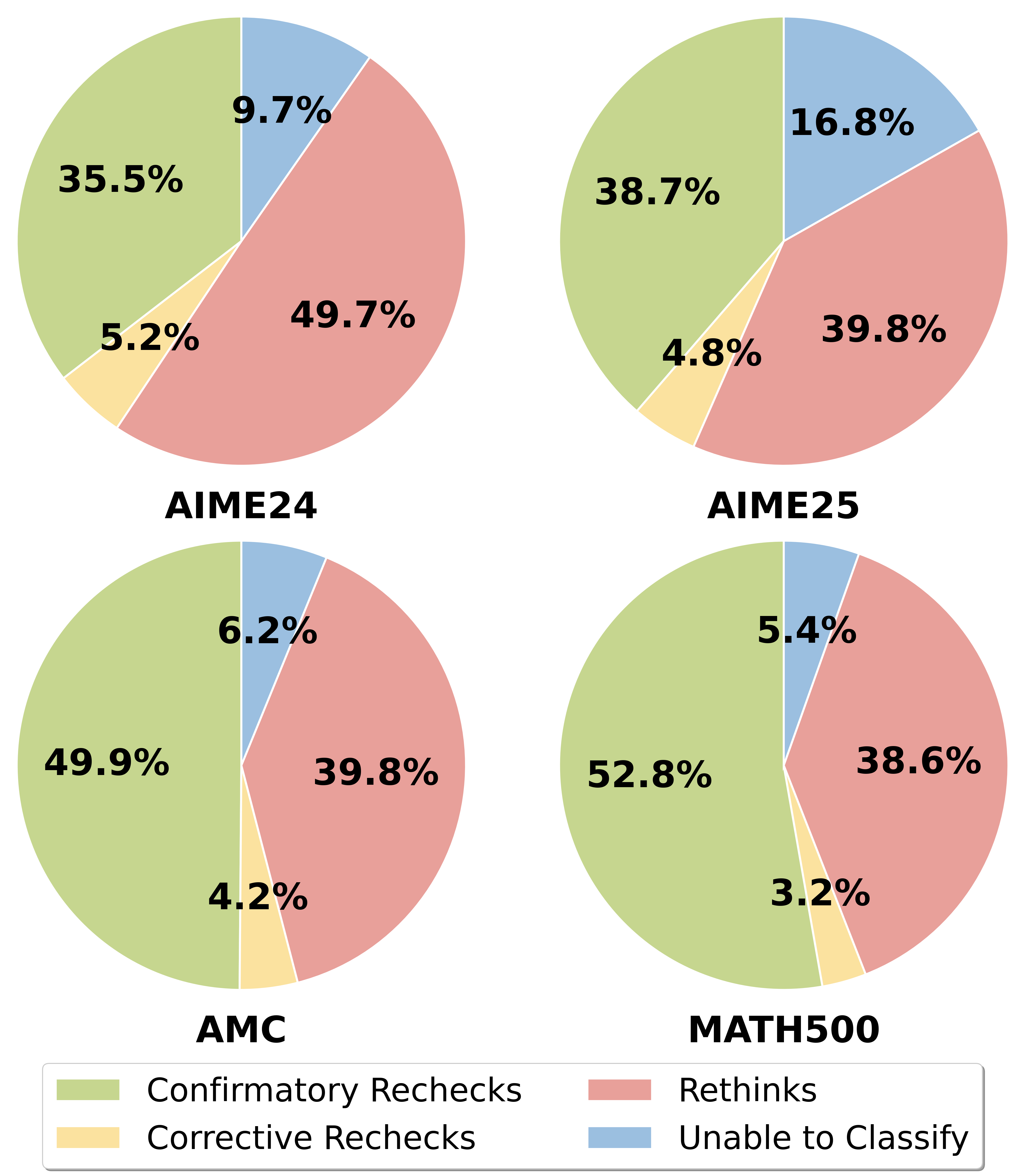}
    \caption{Proportion of reflection types for Qwen3-8B annotated by GPT-5. More results are presented in Appendix \ref{sec:appendixB}. Human evaluation results are presented in Appendix \ref{sec:appendixC}.}
    \label{fig:reflection_proportion}
\end{figure}

\vspace{-0.5em}
\paragraph{Corrective vs. Confirmatory Rechecks.}
We categorize rechecks based on their functional outcome. A corrective recheck identifies an error in a previously stated intermediate result and leads to a correction. In contrast, a confirmatory recheck verifies an earlier step but does not uncover any error and does not alter the mid-result. We prompt GPT-5 to assess whether the recheck reveals an actual mistake and results are also presented in Figure \ref{fig:reflection_proportion}.

\begin{figure*}[t]
	\centering
    \setlength{\belowcaptionskip}{-0.2cm}
	\includegraphics[width=0.9\linewidth]{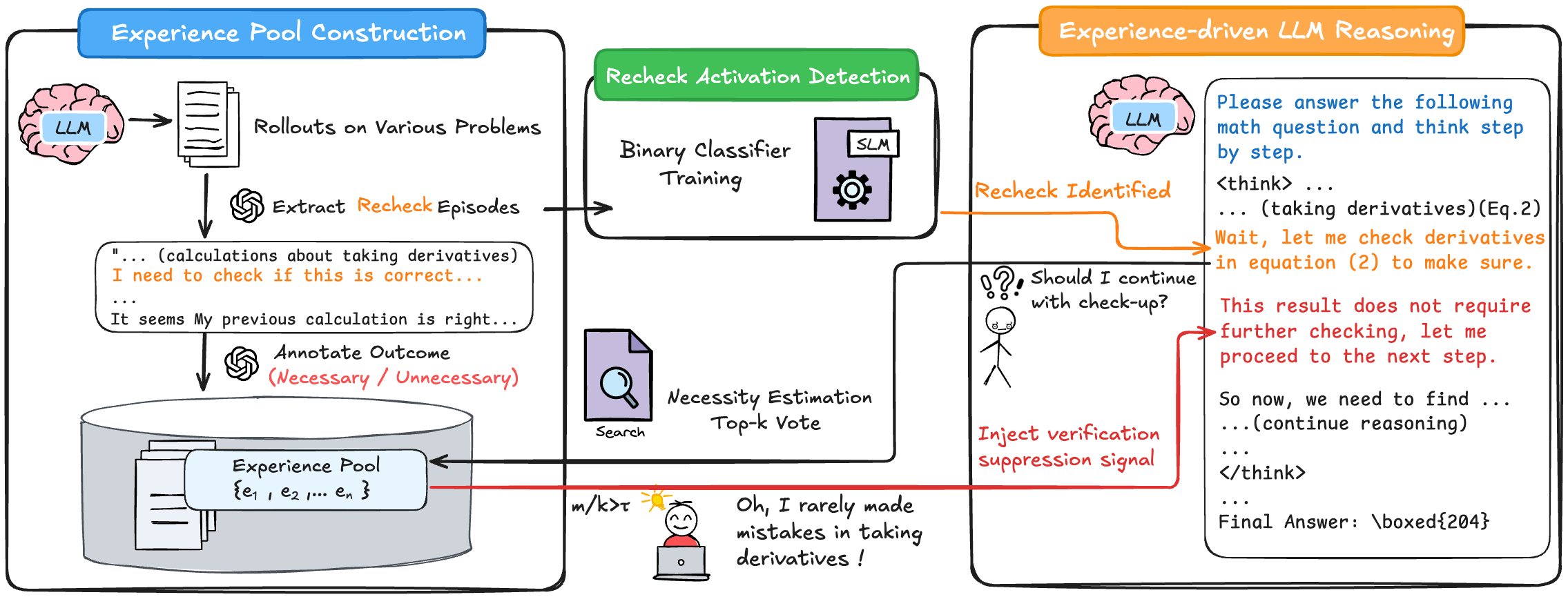}
        \caption{Overview of the experience-driven verification control framework. Recheck episodes are collected and annotated offline to form an experience pool, which captures how often rechecks succeed in correcting errors for specific mathematical manipulations or formula-level derivations, e.g. taking derivatives. The experience pool will be consulted at the test-time to suppress redundant self-verification during LLM reasoning.}
	\label{fig:method}
\end{figure*}

\vspace{-0.5em}
\paragraph{Findings.}
We find that the overwhelming majority of rechecks are confirmatory rather than corrective. Across datasets, approximately 85–95\% of rechecks merely confirm prior reasoning and do not change intermediate results or final answers. And this pattern is consistent across both simpler and harder benchmarks. These results indicate that although models frequently engage in self-verification, such behavior seldom leads to error correction.

All these findings reveal a clear inefficiency: a large number of tokens are evoked to self-verify but almost never change the outcome, the predominance of confirmatory rechecks suggests that self-verification is often overused in current LLM reasoning.



\section{Experience-Guided Self-Verification}
\label{sec:method}

The analysis in the previous section shows that large language models frequently initiate self-verification during reasoning, even in situations where such verification rarely alters intermediate results. As a result, successive self-verification actions often incur additional computational cost while providing limited corrective benefit.

This observation motivates a central question: should every verification attempt be carried out, or can some of them be safely avoided? Intuitively, self-verification becomes redundant when similar reasoning patterns have consistently produced correct results on certain math formulas. Human problem solving offers an analogy. When encountering familiar mathematical forms, such as routine re-computations, constraint checks, or optimality and minimality validations, people often rely on prior experience, reasoning that similar derivations have worked correctly many times before. In such cases, explicit rechecking is often unnecessary.

Inspired by this perspective, we propose an experience-driven self-verification control framework. The core idea is simple: when a model begins to verify a previously stated intermediate result, we consult historical self-verification outcomes from similar contexts and formulas. If past experience indicates that such verification almost never uncovers errors, we discourage the model from continuing the recheck. Otherwise, verification proceeds as usual.

Our approach operates entirely at test time, does not require training or modifying the underlying language model, and introduces minimal inference overhead. While suppressing redundant verification naturally shortens reasoning trajectories and reduces latency, these effects arise as a consequence of avoiding unnecessary verification rather than as an explicit objective.

\subsection{Experience Memory Construction}
\label{sec:pool}
To enable experience-driven suppression of redundant self-verification, we first construct an experience pool that records historical verification outcomes observed in large language model reasoning. This process is illustrated in the left panel of Figure \ref{fig:method}.

\vspace{-0.5em}
\paragraph{Extracting Recheck Activations.}
We begin by collecting reasoning rollouts from a target large language model across a diverse set of problem instances. The same model is used both for experience collection and for inference-time reasoning, ensuring that the extracted verification patterns are representative of the target model’s native behavior. Given a reasoning rollout, we aim to collect experiences where the LLM is performing a recheck. To achieve that, we provide the whole rollout to GPT-5~\citep{singh2025openaigpt5card} to identify and extract all sentences that initiate a recheck of a previously stated intermediate result, such as re-computation, constraint verification, candidate testing, or optimality validation. The set of sentences marks the activation of self-verification behavior. We intentionally focus only on recheck activations and exclude rethink behaviors, such as strategy shifts or reformulations, as rethinking plays a crucial role in correcting reasoning trajectories and should not be suppressed. To improve reliability, we apply additional filtering to ensure that extracted set can match and correspond to explicit verification attempts present in the original rollout. We denote the collection of recheck activation sentences as $\mathcal{S}$.

\vspace{-0.5em}
\paragraph{Outcome Annotation.}
For each identified recheck activation, we extract a self-verification episode, which consists of the local reasoning context and the subsequent verification steps, by taking a fixed-size context window surrounding the activation sentence.
For each extracted self-verification episode, we prompt GPT-5~\citep{singh2025openaigpt5card} again to annotate whether the recheck is necessary or unnecessary, Specifically, 1) \textbf{Necessary}: if the verification identifies an error, or corrects an intermediate result (corrective); 2) \textbf{Unnecessary}, if the verification confirms correctness and does not change the reasoning outcome (confirmatory). Episodes for which GPT-5 cannot make a clear determination are discarded to ensure annotation reliability. 

Each annotated self-verification is stored as an experience unit. Formally, we define the experience pool as
\begin{equation}
    \mathcal{C} = \{ e_1, e_2, \dots, e_n \},
\end{equation}
where each experience unit $e_{i}$ consists of $e_i = (c_i, y_i)$. Here, $c_i$ denotes a textual representation of the self-verification episode context extracted around a recheck activation, and $y_i \in \{0,1\}$ is a binary label indicating whether the recheck is necessary ($y_i = 0$) or unnecessary ($y_i = 1$).

The resulting experience pool $\mathcal{C}$ serves as a reusable memory of historical self-verification behavior, capturing how often rechecks succeed in correcting errors for specific \textbf{local reasoning patterns}, such as concrete mathematical manipulations or formula-level derivations (e.g., taking derivatives of a quadratic equation in Figure \ref{fig:method}).

\begin{table*}[t]
    \caption{Accuracy (Pass@1) and average reasoning length for Base, Full-suppress, and Experience-Driven Suppression (EDS). EDS reduces reasoning length while preserving accuracy relative to the base model. $^\dag$indicates numbers reported in \citet{kang2025trymattersrevisitingrole}.}
    \label{tab:inference_efficiency}
    \centering
    \begin{adjustbox}{width=0.87\linewidth}
    \begin{tabular}{llccc|ccc}
        \toprule[1.20pt]
    \multirow{2}{*}{\textbf{Model}} & \multirow{2}{*}{\textbf{Dataset}} & \multicolumn{3}{c}{\textbf{Accuracy} (\%)} & \multicolumn{3}{c}{\textbf{Length}} \\ 
    \cmidrule(lr){3-8}
                         &    & Base  & Full-supress  & EDS  & Base  & Full-supress & EDS \\ 
                             \midrule
        \multirow{6}{*}{\textbf{Qwen3-8B}} & AIME24    & 74.58 & 70.63 (-3.95) & 72.92 (-1.66) & 14605 & 12734 (-12.8\%) & 13296 (-9.0\%) \\
        & AIME25       & 67.71 & 66.67 (-1.04) & 70.00 (+2.29) & 17133 & 15713 (-8.3\%)  & 16086 (-6.1\%) \\
        & AMC            & 95.62 & 96.25 (+0.63) & 98.75 (+3.13) & 8091  & 6564 (-18.9\%)  & 6893 (-14.8\%) \\
        & Math500        & 95.80 & 95.20 (-0.60) & 97.20 (+1.40) & 4939  & 3935 (-20.3\%)  & 4110 (-16.8\%) \\
        & Olympiad Bench & 80.42 & 79.53 (-0.89) & 79.82 (-0.60) & 10480 & 9540 (-9.0\%)   & 9739 (-7.1\%)  \\
        & Average        & 80.95 & 79.56 (-1.39) & 81.43 (\textbf{+0.48}) & 11361 & 10071 (-11.4\%) & 10386 (-8.6\%) \\
        \cmidrule(lr){3-8}
        & FTM (Average)$^\dag$ & Base & Truncate all & Selective truncate & Base & Truncate all & Selective truncate \\
        & \citep{kang2025trymattersrevisitingrole} & 84.70 & 80.90 (-3.80) & 81.80 (-2.90) & 15,125 & 10,601 (-29.9\%) & 11,414 (-24.5\%) \\
        \midrule
        \multirow{6}{*}{\textbf{DeepSeek-7B}} & AIME24 & 57.50 & 56.67 (-0.83) & 58.75 (+1.25) & 11237 & 10105 (-10.1\%) & 10478 (-6.8\%) \\
        & AIME25       & 39.38 & 35.42 (-3.96) & 36.46 (-2.92) & 12489 & 11221 (-10.1\%) & 11680 (-7.4\%) \\
        & AMC            & 91.25 & 90.00 (-1.25) & 90.63 (-0.62) & 5401  & 5067 (-6.2\%)   & 5145 (-4.7\%)  \\
        & Math500        & 90.60 & 87.20 (-3.40) & 89.80 (-0.80) & 3303  & 2726 (-17.5\%)  & 2891 (-12.5\%) \\
        & Olympiad Bench & 69.00 & 66.91 (-2.09) & 67.95 (-1.05) & 7913  & 7002 (-11.5\%)  & 7183 (-9.2\%)  \\
        & Average        & 66.66 & 64.21 (-2.45) & 65.78 (-0.88) & 8386  & 7467 (-11.0\%)  & 7736 (-7.75\%) \\
        \midrule
        \multirow{6}{*}{\textbf{QWQ-32B}} & AIME2024     & 79.17 & 78.75 (-0.42) & 83.33 (+4.16) & 11237 & 10105 (-13.4\%) & 10478 (-9.5\%) \\
        & AIME2025       & 68.54 & 64.16 (-4.38) & 65.63 (-2.91) & 15811 & 14133 (-10.6\%) & 14908 (-5.7\%) \\
        & AMC            & 97.50 & 93.75 (-3.75) & 95.00 (-2.50) & 7542  & 6526 (-13.5\%)  & 6719 (-10.9\%) \\
        & Math500        & 97.00 & 95.60 (-1.40) & 97.00 (-0.00) & 4659  & 3768 (-19.1\%)  & 3940 (-15.4\%) \\
        & Olympiad Bench & 81.90 & 81.45 (-0.45) & 83.53 (+1.63) & 9602  & 8454 (-12.0\%)  & 8710 (-9.3\%)  \\
        & Average        & 82.91 & 81.21 (-1.70) & 83.48 (\textbf{+0.57}) & 10293 & 8988 (-12.7\%)  & 9379 (-8.9\%) \\
        
        \bottomrule[1.2pt]
    \end{tabular}
    \end{adjustbox}
    \vspace{-10pt}
\end{table*}

\vspace{-0.5em}
\subsection{Experience-Enhanced Inference}
\paragraph{Detecting Recheck Activation}
During inference, a key challenge is to identify when the model is about to start a recheck of a previously stated intermediate result. A natural approach is to instruct the model to explicitly mark such behaviors and emit a special operation when it plans to recheck, analogous to pipelines that interleave internal thoughts with tool calls (e.g., Search-o1~\citep{li2025searcho1agenticsearchenhancedlarge}). However, we find that the model often struggles to reliably follow instructions within the thinking part (inside \texttt{<think>} tags).

Motivated by ~\citet{kang2025trymattersrevisitingrole}, we instead adopt a separate, lightweight identifier that detects whether a generated sentence initiates a recheck and such lightweight components introduce only limited inference latency~\citep{kang2025trymattersrevisitingrole}.
We implement the recheck activation identifier as a binary classifier. Training data is derived from the same reasoning rollouts used for experience pool construction. Specifically, positive examples are given by $\mathcal{S}$ collected in Section~\ref{sec:pool}. Positive examples consist of recheck activation sentences confirmed by GPT-5 during the extraction process described.
Hard negative examples arise naturally during this filtering stage: sentences that are initially extracted as potential recheck activations but subsequently rejected by GPT-5 as not constituting true rechecks are treated as hard negatives.
The trained classifier achieves over 97\% accuracy on the constructed validation and test splits.

\vspace{-0.5em}
\paragraph{Experience Retrieval and Necessity Estimation}
When the identifier predicts that the current sentence initiates a recheck, we construct a query by extracting a fixed-size context window before the recheck activation. This local context typically contains concrete mathematical manipulations or formula-level derivations that characterize the recheck being attempted. We use this context as a query to retrieve the top-$k$ most similar experience units from the experience pool $\mathcal{C}$ using a sparse retriever (BM25~\citep{robertson2029bm25}), which is effective for matching symbolic expressions and mathematical operations while introducing minimal inference-time overhead. Let the retrieved set be $\{e_{i_1}, \dots, e_{i_k}\}$, where each $ e_{i_j} = (c_{i_j}, y_{i_j})$.

To estimate whether the current recheck is likely to be redundant, we do not forward the retrieved candidate texts to the LLM and ask it to decide whether to proceed with rechecking, as doing so would consume additional token budget and introduce new reasoning steps. Instead, we aggregate the outcome labels of the retrieved experience units. Intuitively, if rechecks on similar local reasoning patterns have rarely corrected errors in the past, continuing the current verification is unlikely to change the outcome. Concretely, we estimate the likelihood that a recheck is unnecessary as
\begin{equation}
    \mathcal{P}_{\text{unnec}} = \frac{1}{k} \sum_{j=1}^{k} \mathbb{I}[y_{i_j} = 1],
\end{equation}
where $\mathbb{I}[\cdot]$ is the indicator function. If $ \mathcal{P}_{\text{unnec}}$ exceeds a predefined threshold $\tau$, we treat the current recheck as likely unnecessary.

\vspace{-0.5em}
\paragraph{Experience-Driven Suppression (EDS)}
If a recheck is estimated to be unnecessary, we inject a short verification suppression signal immediately after the activation sentence (e.g., \emph{``This result does not require further checking, let me proceed to the next step.''})\footnote{We observe that the specific wording of the verification suppression signal affects its effectiveness.  Language models exhibit discourse-level attractors for resolving epistemic uncertainty (e.g., phrases such as ``wait'' or ``maybe''), which can cause verification behavior to persist even after an initial interruption. In practice, suppression signals that explicitly assert epistemic closure (e.g., ``is correct'') or provide a forward redirection (e.g., ``proceed'') are more effective at closing verification-related discourse.}. This signal serves as a lightweight directive that discourages the model from continuing the verification process and encourages it to proceed with subsequent reasoning steps.

Empirically, we observe that after receiving such a suppression signal, the model typically skips subsequent re-computation or validation steps and proceeds directly to the next stage of reasoning. Importantly, this intervention targets only recheck behaviors; we do not intervene on rethink behaviors, which are often essential for solving hard problems. As shown in Section~\ref{sec:exp}, suppressing redundant rechecks leads to a consistent reduction in reasoning length, which emerges naturally from avoiding unnecessary self-verification.

\section{Experiments}
\label{sec:exp}

\subsection{Experiment Setups}
We evaluate the effectiveness of Experience-Driven Suppression (EDS) on five mathematical reasoning benchmarks: AIME24 \cite{AIME2024}, AIME25 \cite{AIME2025}, AMC23 \cite{AMC}, MATH500 \cite{hendrycks2021measuring}, and Olympiad Bench \cite{he2024olympiadbench}. Experiments are conducted on three models: Qwen3-8B \cite{yang2025qwen3}, QwQ-32B \cite{qwq32b}, and DeepSeek-R1-Distill-Qwen-7B \cite{deepseekai2025deepseekr1incentivizingreasoningcapability}. 

We compare three inference settings. \textbf{Base}: standard model inference without any intervention. \textbf{Full-suppress}: injecting a suppression signal whenever a recheck activation is identified. \textbf{First Try Matters (FTM)} \citep{kang2025trymattersrevisitingrole}, a test-time method truncates the thinking immediately after the first candidate final answer is produced. \textbf{EDS} applies selective suppression within the thinking process: mid-rechecks are suppressed only when historical experience indicates that the verification is unlikely to be beneficial.


For EDS, we construct the experience pool using reasoning rollouts collected from the Deepscaler dataset~\citep{deepscaler2025}, which consists of AIME and AMC problems prior to 2023. Using GPT-5{\footnote{We use the model \texttt{gpt-5-2025-08-07}.}}~\citep{singh2025openaigpt5card}, we extract recheck activation set $\mathcal{S}$ and annotate whether each recheck is necessary or unnecessary, resulting in an experience pool of 10k units with an unnecessary-to-necessary ratio of approximately 7:3.
Recheck activations during inference are detected using a RoBERTa-base~\citep{Roberta2019}. The classifier is trained on 4,245 instances derived from $\mathcal{S}$, with a positive-to-negative ratio of 1:3 and a balanced split between hard and easy negatives. Easy negatives are uniformly sampled from non-verification sentences in the original rollouts. We train the classifier for 3 epochs with a learning rate of 2e-5, achieving 97\% accuracy on the test split.
During EDS inference, we retrieve the top-30 most similar experience units using BM25~\citep{robertson2029bm25} and suppress a recheck if the proportion of unnecessary experiences exceeds a threshold $\tau = 0.8$. This threshold is set above the base unnecessary rate of the experience pool (0.7).

\subsection{Main Restuls}
Table~\ref{tab:inference_efficiency} reports accuracy and average length. Across all models and datasets, full-suppress consistently shortens reasoning traces, reducing average length by approximately 11–13\%. However, this intervention leads to noticeable accuracy degradation, with average drops ranging from $-1.4$ to $-2.5$ points. 

FTM achieves substantially larger reductions in reasoning length by truncating all content following the first candidate answer. While effective in reducing token usage, this aggressive truncation incurs a pronounced accuracy drop. In particular, for Qwen3-8B, FTM reduces average reasoning length up to 30\% but degrades accuracy by $-3.8$ points. This suggests that their truncation may discard rethink behaviors, which play a crucial role in reasoning and should not be indiscriminately removed.

In contrast, EDS targets redundant recheck behaviors while excluding rethink cases from suppression. EDS achieves consistent reductions in reasoning length while maintaining (in many cases improving) accuracy relative to the base model. For Qwen3-8B and QwQ-32B, EDS improves average accuracy by +0.48 and +0.57 points respectively, while shortening reasoning traces by approximately 9\% on average and up to 20.3\% on MATH500. These results indicate that leveraging historical verification outcomes provides a reliable signal for estimating when a recheck is unlikely to be beneficial, enabling selective suppression without materially affecting correctness. 

Overall, the results indicate that reflections are neither uniformly harmful nor uniformly necessary. 
While truncating reflection wholesale (as in FTM) can substantially reduce token, it risks removing rethink and necessary recheck behaviors. By selectively suppressing rechecks based on prior experience, EDS enables models to bypass unnecessary rechecks while preserving self-verification steps that meaningfully contribute to correctness, achieving a more favorable accuracy–efficiency trade-off.

\subsection{Accuracy–Efficiency Trade-off}
\begin{figure}[t]
    \centering
    \setlength{\belowcaptionskip}{-0.2cm}

    \begin{subfigure}{0.46\columnwidth}
        \centering
        \includegraphics[width=\linewidth]{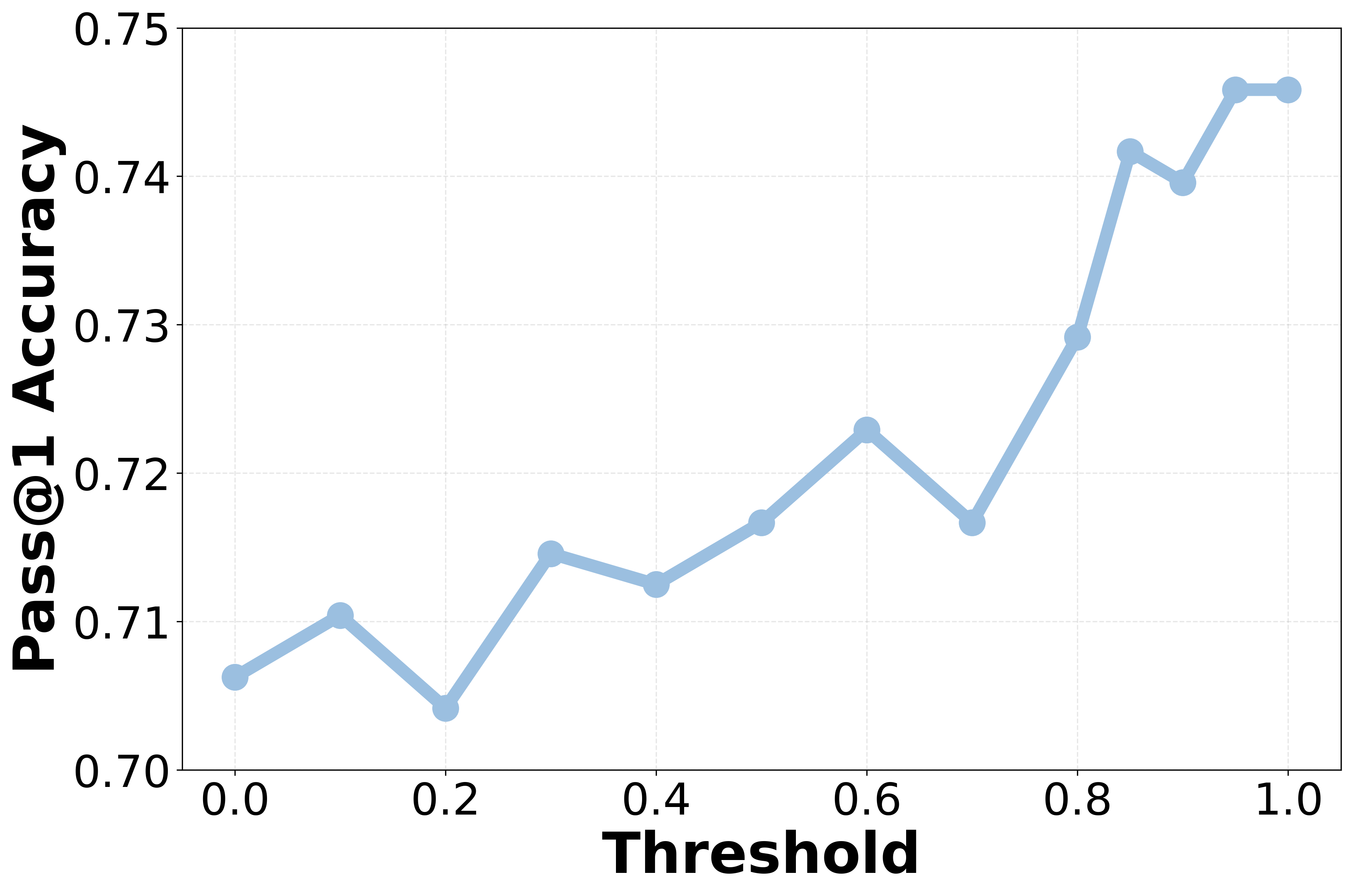}
        \caption{Pass@1}
    \end{subfigure}
    \hfill
    \begin{subfigure}{0.46\columnwidth}
        \centering
        \includegraphics[width=\linewidth]{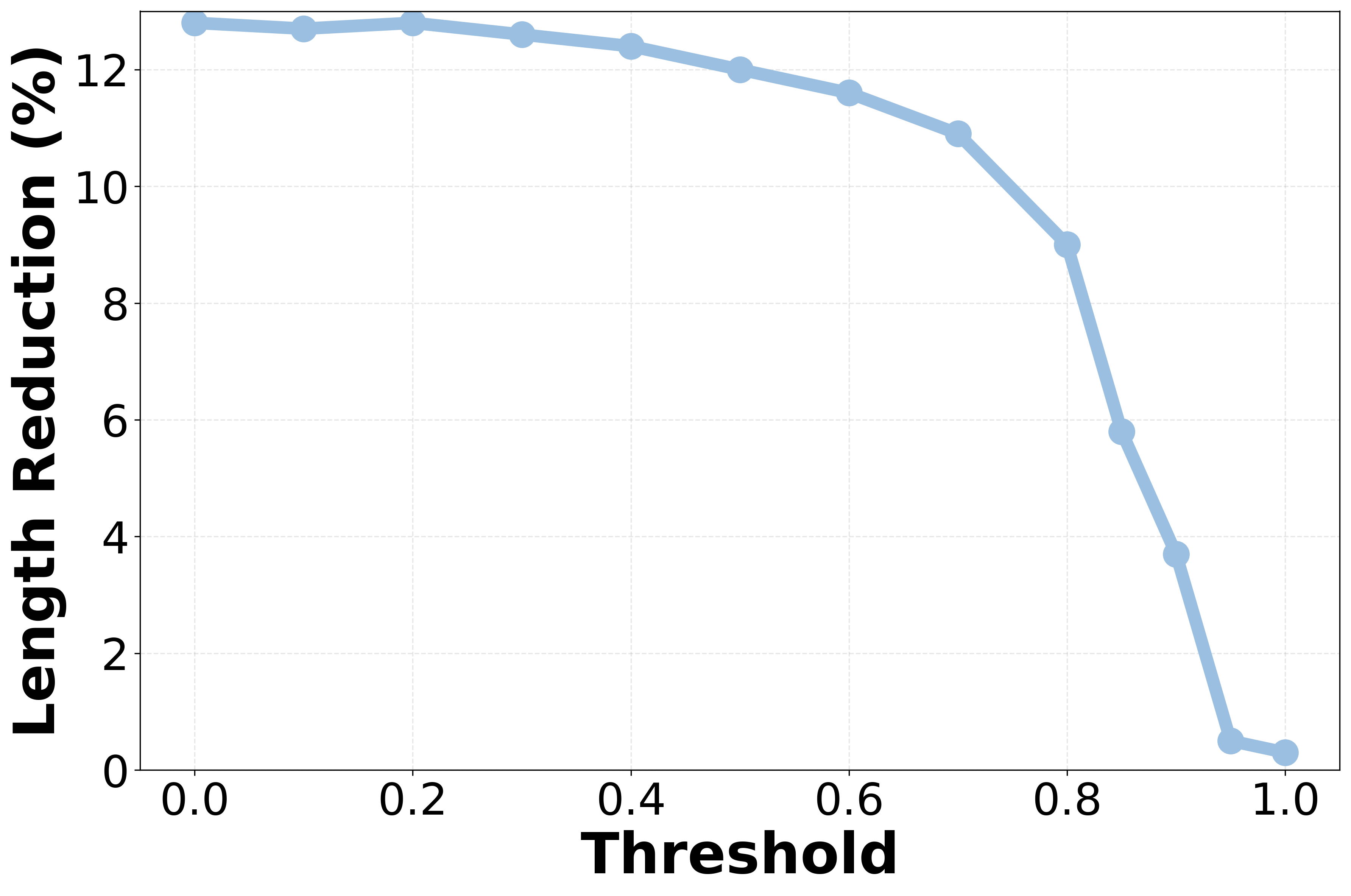}
        \caption{Length reduction (\%)}
    \end{subfigure}
    \caption{Accuracy–efficiency trade-off.}
    \label{fig:exp_train_percentage}
\end{figure}

\begin{figure}[t]
    \centering
    \setlength{\abovecaptionskip}{0.1cm}
    \setlength{\belowcaptionskip}{-0.3cm}
    \includegraphics[width=0.6\linewidth]{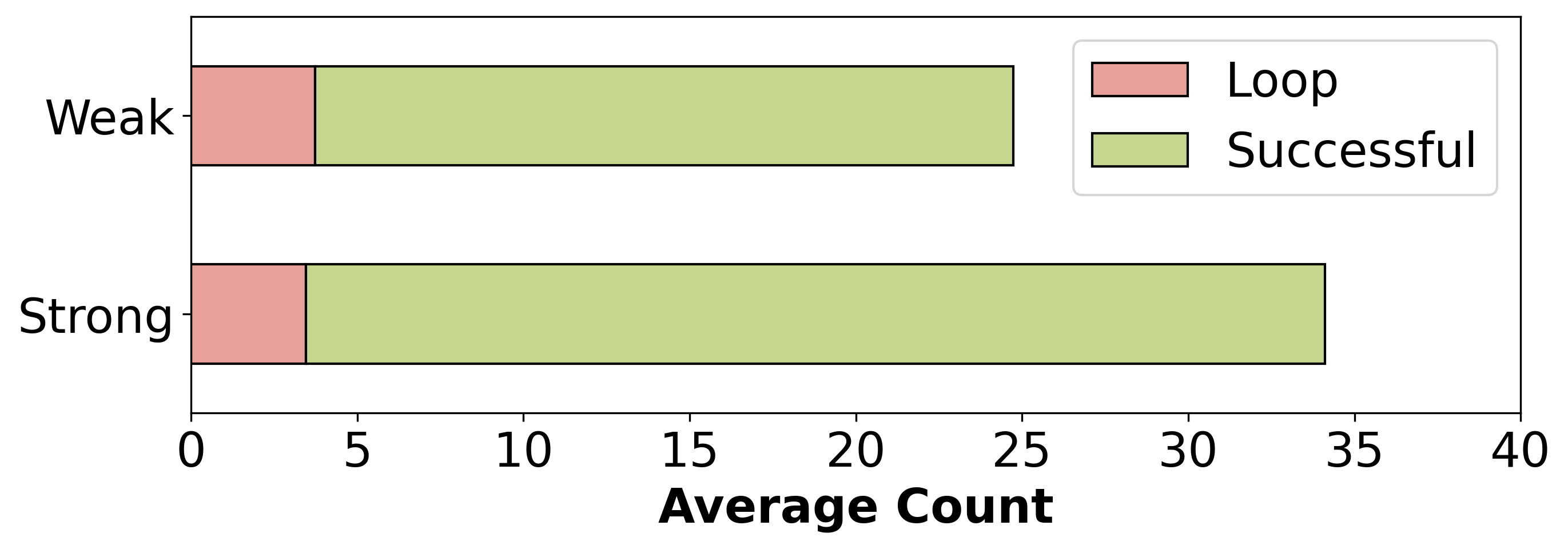}
    \caption{Average recheck activation count per rollout.}
    \label{fig:loop_count}
    \vspace{-7pt}
\end{figure}
Figure \ref{fig:exp_train_percentage} illustrates how the behavior of EDS varies as we adjust the voting threshold $\tau$, which controls how confidently a recheck must be judged unnecessary before suppression is applied. Increasing $\tau$ makes suppression more conservative, resulting in fewer rechecks being filtered.

As shown in Figure \ref{fig:exp_train_percentage}, accuracy increases steadily with larger $\tau$ and saturates around $\tau\approx0.95$, as higher thresholds reduce the likelihood of suppressing rechecks that could be corrective. In contrast, length reduction decreases as $\tau$ increases. In particular, when 
$\tau$ lies in the range $0.7\le\tau\le0.9$, length reduction drops rapidly. 
Together, these trends reveal a clear accuracy–efficiency trade-off: lower thresholds favor stronger suppression and greater reductions in reasoning length, while higher thresholds preserve more verification behavior and yield higher accuracy. This trade-off allows EDS to be flexibly configured according to the desired balance between computational efficiency and reasoning reliability.

\subsection{Case study}

\begin{figure}[t]
    \centering
    \setlength{\abovecaptionskip}{0.1cm}
    \setlength{\belowcaptionskip}{-0.3cm}
    \includegraphics[width=0.9\linewidth]{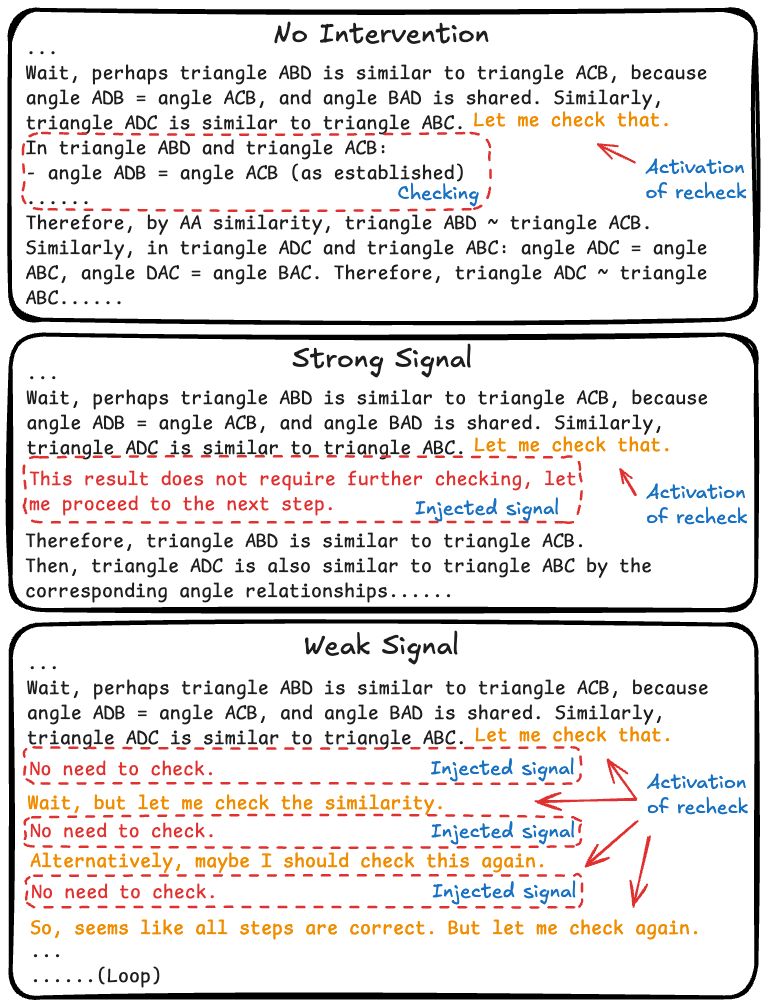}
    \caption{Success and failure cases.}
    \label{fig:case}
\end{figure}

To qualitatively illustrate how verification suppression affects model behavior, Figure~\ref{fig:case} presents representative examples under three settings: no intervention, strong suppression, and weak suppression.

Without intervention, once a recheck is triggered, the model typically proceeds with explicit verification. Injecting a strong suppression signal that asserts epistemic closure and provides forward redirection reliably terminates the verification process. The model accepts the signal as closing the uncertainty and continues with subsequent reasoning. In contrast, weak suppression signals fail to close the verification discourse and may instead provoke repeated recheck attempts, leading to activation loops in which verification is repeatedly re-initiated.

Figure \ref{fig:loop_count} shows that strong signals achieve a higher rate of successful suppression, although looping still occurs. To further stabilize suppression in practice, we introduce a simple cooldown mechanism: after a suppression is applied, recheck detection is temporarily disabled for 3 steps, preventing immediate re-activation of verification.

\section{Conclusion}
\label{sec:conclusion}
We revisit reflection in LRMs through the lens of self-verification, and show it is dominated by repeated rechecks that rarely alter downstream reasoning, exposing a mismatch between verification frequency and utility. Without modifying model parameters, we propose an experience-driven test-time controller that detects recheck activations, retrieves similar past verification episodes from an offline pool, and selectively suppresses low-utility checks while preserving strategy-level rethinking. Our method consistently shortens reasoning while maintaining or slightly improving accuracy across models and benchmarks, suggesting that past verification outcomes can guide when verifications are needed.

\section*{Impact Statement}

This paper presents work whose goal is to advance the field of machine learning by improving the efficiency and controllability of test-time reasoning in LLMs. By reducing redundant self-verification while maintaining solution quality, the proposed approach can lower inference-time compute and latency, enabling more scalable deployment of reliable reasoning models and broader access to such capabilities across resource-constrained settings. We anticipate the primary impacts of this work to be increased energy efficiency and cost-effectiveness of model inference, as well as a clearer methodological foundation for understanding and optimizing reflective behaviors in reasoning.

\nocite{langley00}

\bibliography{example_paper}

\begin{thebibliography}{41}
\providecommand{\natexlab}[1]{#1}
\providecommand{\url}[1]{\texttt{#1}}
\expandafter\ifx\csname urlstyle\endcsname\relax
  \providecommand{\doi}[1]{doi: #1}\else
  \providecommand{\doi}{doi: \begingroup \urlstyle{rm}\Url}\fi

\bibitem[{AIME}(2025)]{AIME2025}
{AIME}.
\newblock {AIME Problems and Solutions, 2025}.
\newblock \url{https://artofproblemsolving.com/wiki/index.php/AIME_Problems_and_Solutions}, 2025.
\newblock Accessed: January 19, 2026.

\bibitem[{AMC12}(2023)]{AMC}
{AMC12}.
\newblock {AMC 12 Problems and Solutions}.
\newblock \url{https://artofproblemsolving.com/wiki/index.php/AMC_12_Problems_and_Solutions}, 2023.
\newblock Accessed: January 19, 2026.

\bibitem[Bercovich et~al.(2025)Bercovich, Levy, Golan, Dabbah, El-Yaniv, Puny, Galil, Moshe, Ronen, Nabwani, Shahaf, Tropp, Karpas, Zilberstein, Zeng, Singhal, Bukharin, Zhang, et~al.]{bercovich2025llamanemotronefficientreasoningmodels}
Bercovich, A., Levy, I., Golan, I., Dabbah, M., El-Yaniv, R., Puny, O., Galil, I., Moshe, Z., Ronen, T., Nabwani, N., Shahaf, I., Tropp, O., Karpas, E., Zilberstein, R., Zeng, J., Singhal, S., Bukharin, A., Zhang, Y., et~al.
\newblock Llama-nemotron: Efficient reasoning models, 2025.
\newblock URL \url{https://arxiv.org/abs/2505.00949}.

\bibitem[DeepSeek-AI et~al.(2025)DeepSeek-AI, Guo, Yang, Zhang, Song, Zhang, Xu, Zhu, Ma, Wang, et~al.]{deepseekai2025deepseekr1incentivizingreasoningcapability}
DeepSeek-AI, Guo, D., Yang, D., Zhang, H., Song, J., Zhang, R., Xu, R., Zhu, Q., Ma, S., Wang, P., et~al.
\newblock Deepseek-r1: Incentivizing reasoning capability in llms via reinforcement learning, 2025.
\newblock URL \url{https://arxiv.org/abs/2501.12948}.

\bibitem[Didolkar et~al.(2025)Didolkar, Ballas, Arora, and Goyal]{anonymous2026metacognitive}
Didolkar, A., Ballas, N., Arora, S., and Goyal, A.
\newblock Metacognitive reuse: Turning recurring llm reasoning into concise behaviors, 2025.
\newblock URL \url{https://arxiv.org/abs/2509.13237}.

\bibitem[Fang et~al.(2025)Fang, Ma, and Wang]{fang2025thinkless}
Fang, G., Ma, X., and Wang, X.
\newblock Thinkless: Llm learns when to think.
\newblock \emph{Advances in neural information processing systems}, 2025.

\bibitem[He et~al.(2024)He, Luo, Bai, Hu, Thai, Shen, Hu, Han, Huang, Zhang, et~al.]{he2024olympiadbench}
He, C., Luo, R., Bai, Y., Hu, S., Thai, Z.~L., Shen, J., Hu, J., Han, X., Huang, Y., Zhang, Y., et~al.
\newblock Olympiadbench: A challenging benchmark for promoting agi with olympiad-level bilingual multimodal scientific problems.
\newblock \emph{arXiv preprint arXiv:2402.14008}, 2024.
\newblock URL \url{https://aclanthology.org/2024.acl-long.211/}.

\bibitem[Hendrycks et~al.(2021)Hendrycks, Burns, Kadavath, Arora, Basart, Tang, Song, and Steinhardt]{hendrycks2021measuring}
Hendrycks, D., Burns, C., Kadavath, S., Arora, A., Basart, S., Tang, E., Song, D., and Steinhardt, J.
\newblock Measuring mathematical problem solving with the math dataset.
\newblock \emph{arXiv preprint arXiv:2103.03874}, 2021.
\newblock URL \url{https://arxiv.org/abs/2103.03874}.

\bibitem[Hu et~al.(2025)Hu, Chen, Chen, Mu, Shao, and Luo]{hu-etal-2025-hiagent}
Hu, M., Chen, T., Chen, Q., Mu, Y., Shao, W., and Luo, P.
\newblock {H}i{A}gent: Hierarchical working memory management for solving long-horizon agent tasks with large language model.
\newblock In Che, W., Nabende, J., Shutova, E., and Pilehvar, M.~T. (eds.), \emph{Proceedings of the 63rd Annual Meeting of the Association for Computational Linguistics (Volume 1: Long Papers)}, pp.\  32779--32798, Vienna, Austria, July 2025. Association for Computational Linguistics.
\newblock ISBN 979-8-89176-251-0.
\newblock \doi{10.18653/v1/2025.acl-long.1575}.
\newblock URL \url{https://aclanthology.org/2025.acl-long.1575/}.

\bibitem[Jiang et~al.(2025)Jiang, Wu, Huang, Dong, Chi, Dong, Zhang, Lv, Cui, and Wei]{jiang2025think}
Jiang, L., Wu, X., Huang, S., Dong, Q., Chi, Z., Dong, L., Zhang, X., Lv, T., Cui, L., and Wei, F.
\newblock Think only when you need with large hybrid-reasoning models.
\newblock In \emph{The Thirty-ninth Annual Conference on Neural Information Processing Systems}, 2025.
\newblock URL \url{https://openreview.net/forum?id=fDjDVE4qdj}.

\bibitem[Kang et~al.(2025)Kang, Deng, Xiao, Mo, Lee, and Bing]{kang2025trymattersrevisitingrole}
Kang, L., Deng, Y., Xiao, Y., Mo, Z., Lee, W.~S., and Bing, L.
\newblock First try matters: Revisiting the role of reflection in reasoning models, 2025.
\newblock URL \url{https://arxiv.org/abs/2510.08308}.

\bibitem[Langley(2000)]{langley00}
Langley, P.
\newblock Crafting papers on machine learning.
\newblock In Langley, P. (ed.), \emph{Proceedings of the 17th International Conference on Machine Learning (ICML 2000)}, pp.\  1207--1216, Stanford, CA, 2000. Morgan Kaufmann.

\bibitem[Li et~al.(2025)Li, Dong, Jin, Zhang, Zhou, Zhu, Zhang, and Dou]{li2025searcho1agenticsearchenhancedlarge}
Li, X., Dong, G., Jin, J., Zhang, Y., Zhou, Y., Zhu, Y., Zhang, P., and Dou, Z.
\newblock Search-o1: Agentic search-enhanced large reasoning models, 2025.
\newblock URL \url{https://arxiv.org/abs/2501.05366}.

\bibitem[Liu \& Wang(2025)Liu and Wang]{liu-wang-2025-answer}
Liu, X. and Wang, L.
\newblock Answer convergence as a signal for early stopping in reasoning.
\newblock In Christodoulopoulos, C., Chakraborty, T., Rose, C., and Peng, V. (eds.), \emph{Proceedings of the 2025 Conference on Empirical Methods in Natural Language Processing}, pp.\  17896--17907, Suzhou, China, November 2025. Association for Computational Linguistics.
\newblock ISBN 979-8-89176-332-6.
\newblock \doi{10.18653/v1/2025.emnlp-main.904}.
\newblock URL \url{https://aclanthology.org/2025.emnlp-main.904/}.

\bibitem[Liu et~al.(2019)Liu, Ott, Goyal, Du, Joshi, Chen, Levy, Lewis, Zettlemoyer, and Stoyanov]{Roberta2019}
Liu, Y., Ott, M., Goyal, N., Du, J., Joshi, M., Chen, D., Levy, O., Lewis, M., Zettlemoyer, L., and Stoyanov, V.
\newblock Roberta: {A} robustly optimized {BERT} pretraining approach.
\newblock \emph{CoRR}, abs/1907.11692, 2019.
\newblock URL \url{http://arxiv.org/abs/1907.11692}.

\bibitem[Liu et~al.(2025{\natexlab{a}})Liu, Si, Narasimhan, and Yao]{liu-etal-2025-contextual}
Liu, Y., Si, C., Narasimhan, K.~R., and Yao, S.
\newblock Contextual experience replay for self-improvement of language agents.
\newblock In Che, W., Nabende, J., Shutova, E., and Pilehvar, M.~T. (eds.), \emph{Proceedings of the 63rd Annual Meeting of the Association for Computational Linguistics (Volume 1: Long Papers)}, pp.\  14179--14198, Vienna, Austria, July 2025{\natexlab{a}}. Association for Computational Linguistics.
\newblock ISBN 979-8-89176-251-0.
\newblock \doi{10.18653/v1/2025.acl-long.694}.
\newblock URL \url{https://aclanthology.org/2025.acl-long.694/}.

\bibitem[Liu et~al.(2025{\natexlab{b}})Liu, Chen, Li, Pang, Du, and Lin]{liu2025oatzero}
Liu, Z., Chen, C., Li, W., Pang, T., Du, C., and Lin, M.
\newblock There may not be aha moment in r1-zero-like training — a pilot study.
\newblock \url{[https://oatllm.notion.site/oat-zero](https://oatllm.notion.site/oat-zero)}, 2025{\natexlab{b}}.
\newblock Notion Blog.

\bibitem[Luo et~al.(2025)Luo, Tan, Wong, Shi, Tang, Roongta, Cai, Luo, Zhang, Li, Popa, and Stoica]{deepscaler2025}
Luo, M., Tan, S., Wong, J., Shi, X., Tang, W., Roongta, M., Cai, C., Luo, J., Zhang, T., Li, E., Popa, R.~A., and Stoica, I.
\newblock Deepscaler: Surpassing o1-preview with a 1.5b model by scaling rl.
\newblock \url{https://pretty-radio-b75.notion.site/DeepScaleR-Surpassing-O1-Preview-with-a-1-5B-Model-by-Scaling-RL-19681902c1468005bed8ca303013a4e2}, 2025.
\newblock Notion Blog.

\bibitem[{MAA}(2024)]{AIME2024}
{MAA}.
\newblock {American Invitational Mathematics Examination (AIME) 2024}.
\newblock {\url{https://maa.org/math-competitions/american-invitational-mathematics-examination-aime}}, February 2024.
\newblock Accessed: January 19, 2026.

\bibitem[Mao et~al.(2025)Mao, Yin, Zhu, and Fang]{anonymous2026early}
Mao, M., Yin, B., Zhu, Y., and Fang, X.
\newblock Early stopping chain-of-thoughts in large language models, 2025.
\newblock URL \url{https://arxiv.org/abs/2509.14004}.

\bibitem[Ning et~al.(2025)Ning, Li, Fang, Tan, and Liu]{anonymous2026not}
Ning, Y., Li, W., Fang, J., Tan, N., and Liu, H.
\newblock Not all thoughts are generated equal: Efficient llm reasoning via multi-turn reinforcement learning, 2025.
\newblock URL \url{https://arxiv.org/abs/2505.11827}.

\bibitem[OpenAI(2025)]{openai2025gptoss120bgptoss20bmodel}
OpenAI.
\newblock gpt-oss-120b \& gpt-oss-20b model card, 2025.
\newblock URL \url{https://arxiv.org/abs/2508.10925}.

\bibitem[OpenAI et~al.(2024)OpenAI, :, Jaech, Kalai, Lerer, Richardson, El-Kishky, Low, Helyar, Madry, Beutel, Carney, Iftimie, Karpenko, Passos, Neitz, Prokofiev, Wei, Tam, Bennett, Kumar, Saraiva, Vallone, Duberstein, Kondrich, Mishchenko, et~al.]{openai2024openaio1card}
OpenAI, :, Jaech, A., Kalai, A., Lerer, A., Richardson, A., El-Kishky, A., Low, A., Helyar, A., Madry, A., Beutel, A., Carney, A., Iftimie, A., Karpenko, A., Passos, A.~T., Neitz, A., Prokofiev, A., Wei, A., Tam, A., Bennett, A., Kumar, A., Saraiva, A., Vallone, A., Duberstein, A., Kondrich, A., Mishchenko, A., et~al.
\newblock Openai o1 system card, 2024.
\newblock URL \url{https://arxiv.org/abs/2412.16720}.

\bibitem[OpenAI et~al.(2025)OpenAI, :, Singh, Fry, Perelman, Tart, Ganesh, El-Kishky, McLaughlin, Low, Ostrow, Ananthram, Nathan, Luo, Helyar, Madry, Efremov, Spyra, Baker-Whitcomb, Beutel, Karpenko, Makelov, Neitz, Wei, Barr, Kirchmeyer, Ivanov, Christakis, Gillespie, Tam, Bennett, Wan, Huang, Sandjideh, Yang, et~al.]{singh2025openaigpt5card}
OpenAI, :, Singh, A., Fry, A., Perelman, A., Tart, A., Ganesh, A., El-Kishky, A., McLaughlin, A., Low, A., Ostrow, A., Ananthram, A., Nathan, A., Luo, A., Helyar, A., Madry, A., Efremov, A., Spyra, A., Baker-Whitcomb, A., Beutel, A., Karpenko, A., Makelov, A., Neitz, A., Wei, A., Barr, A., Kirchmeyer, A., Ivanov, A., Christakis, A., Gillespie, A., Tam, A., Bennett, A., Wan, A., Huang, A., Sandjideh, A.~M., Yang, A., et~al.
\newblock Openai gpt-5 system card, 2025.
\newblock URL \url{https://arxiv.org/abs/2601.03267}.

\bibitem[Robertson \& Zaragoza(2009)Robertson and Zaragoza]{robertson2029bm25}
Robertson, S.~E. and Zaragoza, H.
\newblock The probabilistic relevance framework: {BM25} and beyond.
\newblock \emph{Found. Trends Inf. Retr.}, pp.\  333--389, 2009.

\bibitem[Shen et~al.(2025)Shen, Zhang, Huang, Shi, Zhang, Yan, Wang, Wang, Liu, and Lian]{shen-etal-2025-dast}
Shen, Y., Zhang, J., Huang, J., Shi, S., Zhang, W., Yan, J., Wang, N., Wang, K., Liu, Z., and Lian, S.
\newblock {DAST}: Difficulty-adaptive slow-thinking for large reasoning models.
\newblock In Potdar, S., Rojas-Barahona, L., and Montella, S. (eds.), \emph{Proceedings of the 2025 Conference on Empirical Methods in Natural Language Processing: Industry Track}, pp.\  2322--2331, Suzhou (China), November 2025. Association for Computational Linguistics.
\newblock ISBN 979-8-89176-333-3.
\newblock \doi{10.18653/v1/2025.emnlp-industry.160}.
\newblock URL \url{https://aclanthology.org/2025.emnlp-industry.160/}.

\bibitem[Sui et~al.(2025)Sui, Chuang, Wang, Zhang, Zhang, Yuan, Liu, Wen, Zhong, Chen, and Hu]{sui2025stopoverthinkingsurveyefficient}
Sui, Y., Chuang, Y.-N., Wang, G., Zhang, J., Zhang, T., Yuan, J., Liu, H., Wen, A., Zhong, S., Chen, H., and Hu, X.
\newblock Stop overthinking: A survey on efficient reasoning for large language models, 2025.
\newblock URL \url{https://arxiv.org/abs/2503.16419}.

\bibitem[Team(2025)]{qwq32b}
Team, Q.
\newblock Qwq-32b: Embracing the power of reinforcement learning, March 2025.
\newblock URL \url{https://qwenlm.github.io/blog/qwq-32b/}.

\bibitem[Wang et~al.(2025{\natexlab{a}})Wang, Guo, Ma, and Zhang]{wang-etal-2025-far}
Wang, J., Guo, Z., Ma, W., and Zhang, M.
\newblock How far can {LLM}s improve from experience? measuring test-time learning ability in {LLM}s with human comparison.
\newblock In Christodoulopoulos, C., Chakraborty, T., Rose, C., and Peng, V. (eds.), \emph{Proceedings of the 2025 Conference on Empirical Methods in Natural Language Processing}, pp.\  25677--25691, Suzhou, China, November 2025{\natexlab{a}}. Association for Computational Linguistics.
\newblock ISBN 979-8-89176-332-6.
\newblock \doi{10.18653/v1/2025.emnlp-main.1304}.
\newblock URL \url{https://aclanthology.org/2025.emnlp-main.1304/}.

\bibitem[Wang et~al.(2025{\natexlab{b}})Wang, Liu, Zhang, Chen, and Huang]{wang2025patsprocessleveladaptivethinking}
Wang, Y., Liu, J., Zhang, S., Chen, J., and Huang, S.
\newblock Pats: Process-level adaptive thinking mode switching, 2025{\natexlab{b}}.
\newblock URL \url{https://arxiv.org/abs/2505.19250}.

\bibitem[Wu et~al.(2025)Wu, Xie, Zhang, Chen, Zhang, Su, and Xiao]{wu2025arm}
Wu, S., Xie, J., Zhang, Y., Chen, A., Zhang, K., Su, Y., and Xiao, Y.
\newblock {ARM}: Adaptive reasoning model.
\newblock In \emph{The Thirty-ninth Annual Conference on Neural Information Processing Systems}, 2025.
\newblock URL \url{https://openreview.net/forum?id=z9oeQrcNh9}.

\bibitem[Yang et~al.(2025{\natexlab{a}})Yang, Li, Yang, Zhang, Hui, Zheng, Yu, Gao, Huang, Lv, et~al.]{yang2025qwen3}
Yang, A., Li, A., Yang, B., Zhang, B., Hui, B., Zheng, B., Yu, B., Gao, C., Huang, C., Lv, C., et~al.
\newblock Qwen3 technical report.
\newblock \emph{arXiv preprint arXiv:2505.09388}, 2025{\natexlab{a}}.
\newblock URL \url{https://arxiv.org/abs/2505.09388}.

\bibitem[Yang et~al.(2025{\natexlab{b}})Yang, Wu, Chen, Xiao, Yang, Wong, and Wang]{yang2025understandingahamomentsexternal}
Yang, S., Wu, J., Chen, X., Xiao, Y., Yang, X., Wong, D.~F., and Wang, D.
\newblock Understanding aha moments: from external observations to internal mechanisms, 2025{\natexlab{b}}.
\newblock URL \url{https://arxiv.org/abs/2504.02956}.

\bibitem[Yufei~Xiang \& Nguyen(2024)Yufei~Xiang and Nguyen]{yufei2024retrospex}
Yufei~Xiang, Yiqun~Shen, Y.~Z. and Nguyen, C.-T.
\newblock Retrospex: Language agent meets offline reinforcement learning critic.
\newblock In \emph{Proceedings of the 2024 Conference on Empirical Methods in Natural Language Processing, {EMNLP}}, 2024.

\bibitem[Zeng et~al.(2025)Zeng, Huang, Li, Zhang, and Deng]{anonymous2026done}
Zeng, Z., Huang, X., Li, B., Zhang, H., and Deng, Z.
\newblock Done is better than perfect: Unlocking efficient reasoning by structured multi-turn decomposition, 2025.
\newblock URL \url{https://arxiv.org/abs/2505.19788}.

\bibitem[Zhai et~al.(2025)Zhai, Tao, Chen, Zou, Chen, Fu, Mai, Yu, Deng, Cao, Liu, Ding, and Zhou]{AgentEvolver2025}
Zhai, Y., Tao, S., Chen, C., Zou, A., Chen, Z., Fu, Q., Mai, S., Yu, L., Deng, J., Cao, Z., Liu, Z., Ding, B., and Zhou, J.
\newblock Agentevolver: Towards efficient self-evolving agent system, 2025.
\newblock URL \url{https://arxiv.org/abs/2511.10395}.

\bibitem[Zhang et~al.(2025{\natexlab{a}})Zhang, Lin, Hou, Feng, and Li]{zhang-etal-2025-adaptthink}
Zhang, J., Lin, N., Hou, L., Feng, L., and Li, J.
\newblock {A}dapt{T}hink: Reasoning models can learn when to think.
\newblock In Christodoulopoulos, C., Chakraborty, T., Rose, C., and Peng, V. (eds.), \emph{Proceedings of the 2025 Conference on Empirical Methods in Natural Language Processing}, pp.\  3716--3730, Suzhou, China, November 2025{\natexlab{a}}. Association for Computational Linguistics.
\newblock ISBN 979-8-89176-332-6.
\newblock \doi{10.18653/v1/2025.emnlp-main.184}.
\newblock URL \url{https://aclanthology.org/2025.emnlp-main.184/}.

\bibitem[Zhang et~al.(2026)Zhang, Wu, Chen, Zhang, Li, Lou, Zhou, Zhou, Wang, and Wang]{zhang2026othinkr1intrinsicfastslowthinking}
Zhang, S., Wu, J., Chen, J., Zhang, C., Li, Z., Lou, X., Zhou, W., Zhou, S., Wang, C., and Wang, J.
\newblock Othink-r1: Intrinsic fast/slow thinking mode switching for over-reasoning mitigation, 2026.
\newblock URL \url{https://arxiv.org/abs/2506.02397}.

\bibitem[Zhang et~al.(2025{\natexlab{b}})Zhang, Ruan, Ma, Zhu, Zhao, Li, Chen, Zeng, and Cai]{zhang-etal-2025-continue}
Zhang, X., Ruan, J., Ma, X., Zhu, Y., Zhao, H., Li, H., Chen, J., Zeng, K., and Cai, X.
\newblock When to continue thinking: Adaptive thinking mode switching for efficient reasoning.
\newblock In Christodoulopoulos, C., Chakraborty, T., Rose, C., and Peng, V. (eds.), \emph{Findings of the Association for Computational Linguistics: EMNLP 2025}, pp.\  5808--5828, Suzhou, China, November 2025{\natexlab{b}}. Association for Computational Linguistics.
\newblock ISBN 979-8-89176-335-7.
\newblock \doi{10.18653/v1/2025.findings-emnlp.310}.
\newblock URL \url{https://aclanthology.org/2025.findings-emnlp.310/}.

\bibitem[Zhao et~al.(2024)Zhao, Huang, Xu, Lin, Liu, and Huang]{zhao2024expel}
Zhao, A., Huang, D., Xu, Q., Lin, M., Liu, Y.-J., and Huang, G.
\newblock Expel: Llm agents are experiential learners.
\newblock In Wooldridge, M.~J., Dy, J.~G., and Natarajan, S. (eds.), \emph{Thirty-Eighth {AAAI} Conference on Artificial Intelligence, {AAAI} 2024, Thirty-Sixth Conference on Innovative Applications of Artificial Intelligence, {IAAI} 2024, Fourteenth Symposium on Educational Advances in Artificial Intelligence, {EAAI} 2024, February 20-27, 2024, Vancouver, Canada}, pp.\  19632--19642. {AAAI} Press, 2024.
\newblock \doi{10.1609/aaai.v38i17.29936}.
\newblock URL \url{https://ojs.aaai.org/index.php/AAAI/article/view/29936}.

\bibitem[Zhao et~al.(2025)Zhao, Yan, Shen, Xu, Zhang, Song, Shao, Lu, Xiao, and Zhuang]{zhao2025let}
Zhao, H., Yan, Y., Shen, Y., Xu, H., Zhang, W., Song, K., Shao, J., Lu, W., Xiao, J., and Zhuang, Y.
\newblock Let {LRM}s break free from overthinking via self-braking tuning.
\newblock In \emph{The Thirty-ninth Annual Conference on Neural Information Processing Systems}, 2025.
\newblock URL \url{https://openreview.net/forum?id=u3a2AX0icx}.

\end{thebibliography}
\bibliographystyle{icml2026}

\newpage
\appendix
\onecolumn

\section{Benchmark Statistics}
\label{sec:appendixA}
\begin{table}[h]
\centering
\vspace{10mm}
\caption{Statistics of Datasets and Rollouts}
\label{tab:dataset_stats}
\begin{tabular}{lccc}
\toprule[1.2pt]
Dataset & Problems & Rollouts per Problem & Total Rollouts \\
\midrule
AIME24 & 30 & 16 & 480 \\
AIME25 & 30 & 16 & 480 \\
AMC23 & 40 & 4 & 160 \\
MATH500 & 500 & 1 & 500 \\
Olympiad Bench & 674 & 1 & 674 \\
\midrule
Total & 1317 & - & 2294 \\
\bottomrule[1.2pt]
\end{tabular}
\end{table}

\section{Reflection Type Statistics}
\label{sec:appendixB}
\begin{figure}[!htbp]
    \centering
    \includegraphics[width=0.9\linewidth]{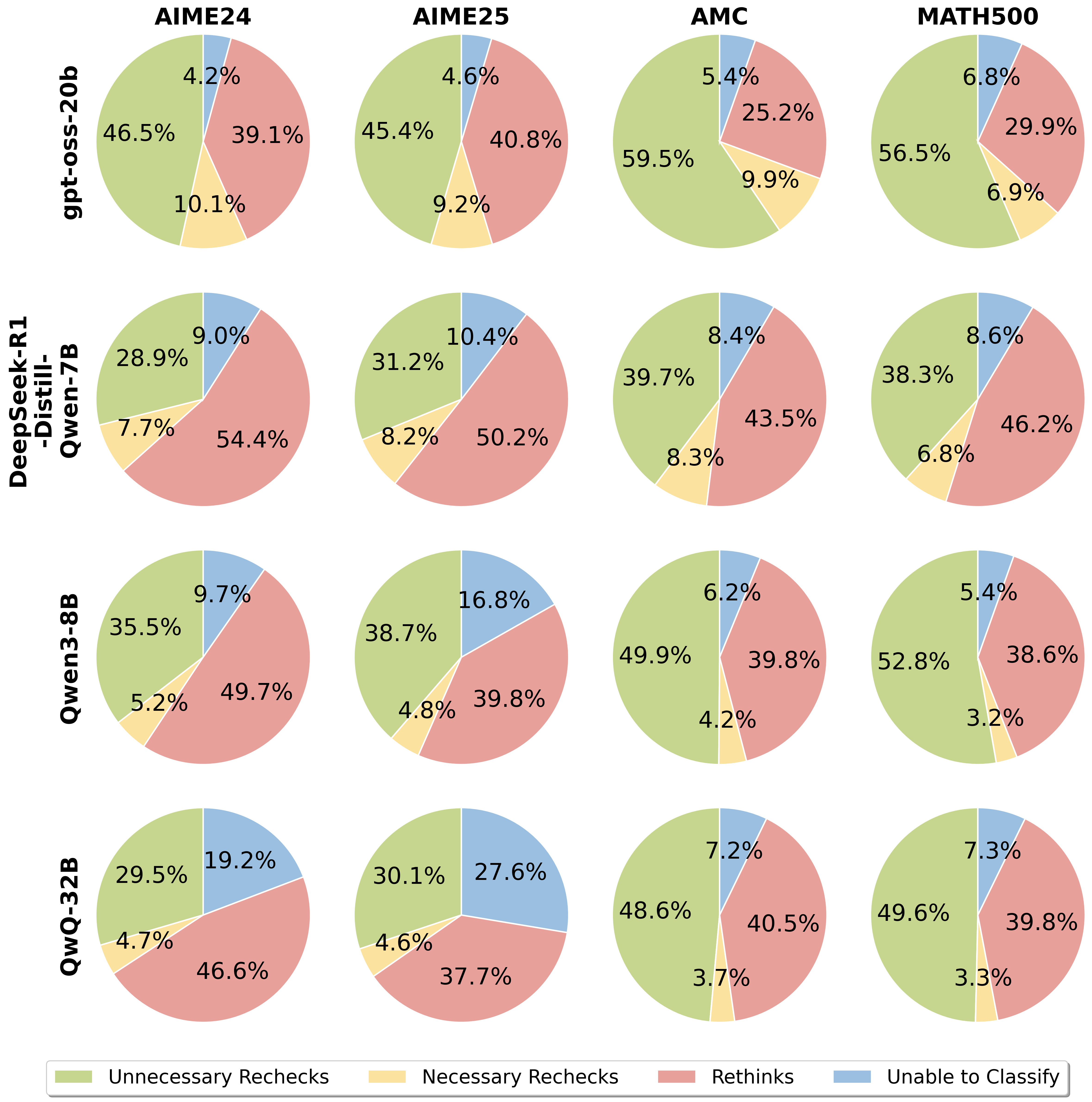}
    \caption{Proportion of reflection types for Qwen3-8B, QwQ-32B, Deepseek-R1-Distilled-Qwen-7B and gpt-oss-20b annotated by GPT-5}
    \label{fig:reflection_all_models}
\end{figure}

To investigate the semantic composition of reflection steps, we analyze four recent models—Qwen3-8B \cite{yang2025qwen3}, QwQ-32B \cite{qwq32b}, gpt-oss-20b \cite{openai2025gptoss120bgptoss20bmodel}, and DeepSeek-R1-Distill-Qwen-7B \cite{deepseekai2025deepseekr1incentivizingreasoningcapability}—across a diverse set of benchmarks, including AIME24 \cite{AIME2024}, AIME25 \cite{AIME2025}, AMC \cite{AMC}, and MATH500 \cite{hendrycks2021measuring}. This setup allows us to examine how reflection behaviors vary across model scales and problem domains.

From Figure \ref{fig:reflection_all_models}, we observe a persistent inefficiency in self-verification behaviors across models and benchmarks. Specifically, the ratio of unnecessary rechecks to necessary rechecks remains consistently high. This pattern indicates a fundamental limitation of current chain-of-thought (CoT) paradigms: models exhibit difficulty in discerning when verification is warranted, leading to substantial computational overhead due to redundant verification steps.

\section{Human Evaluation}
\label{sec:appendixC}
To evaluate the reliability of the LLM-based reflection classifier, we conducted a human evaluation on a random subset of 1,000 reflection steps. Each step was independently annotated by a human expert following the same reflection schema. We compared LLM and human annotations in terms of label distributions and computed both overall agreement and Cohen’s $\kappa$ to quantify alignment beyond chance.

\begin{table}[h]
\centering
\begin{tabular}{lcc}
\toprule
\textbf{Label} & \textbf{Human} & \textbf{LLM} \\
\midrule
Corrective Rechecks & 54 (5.4\%)  & 48 (4.8\%) \\
Confirmatory Rechecks & 428 (42.8\%) & 374 (37.4\%) \\
Rethink & 518 (51.8\%) & 489 (48.9\%) \\
Unable to Classify & - & 89 (8.9\%) \\
\bottomrule
\end{tabular}
\caption{Distribution of reflection types in human and LLM annotations ($n = 1000$).}
\label{tab:reflection_distribution}
\end{table}

Table~\ref{tab:reflection_distribution} shows that the LLM’s predicted label distribution closely matches that of the human annotator across the primary reflection categories, with minor discrepancies attributable to the LLM’s additional “Unable to Classify” label. The confusion matrix in 

\begin{table}[h]
\centering
\begin{tabular}{l|cccc}
\toprule
\textbf{Human $\backslash$ LLM} & \textbf{Corrective Rechecks} & \textbf{Confirmatory Rechecks} & \textbf{Rethinks} & \textbf{Unable to Classify} \\
\midrule
\textbf{Corrective Rechecks} & 47 & 1 & 1 & 5 \\
\textbf{Confirmatory Rechecks} & 1 & 369 & 17 & 41 \\
\textbf{Rethinks} & 0 & 4 & 471 & 43 \\
\textbf{Unable to Classify} & 0 & 0 & 0 & 0 \\
\bottomrule
\end{tabular}
\caption{Confusion matrix between human annotations (rows) and LLM predictions (columns) over 1,000 reflection steps.}
\label{tab:reflection_confusion}
\end{table}

Table~\ref{tab:reflection_confusion} further demonstrates that most disagreements occur between conceptually adjacent categories (e.g., corrective vs. confirmatory rechecks), indicating that misclassifications primarily arise from boundary ambiguity rather than systematic errors.

Quantitatively, the LLM achieves an overall agreement of 88.7\% with human annotations and a Cohen’s $\kappa$ of 0.81, suggesting strong concordance beyond chance. Notably, the model rarely confuses non-adjacent reflection types, reinforcing the semantic coherence of its predictions.

These results indicate that the LLM provides an accurate and reliable approximation of expert judgment for reflection classification, supporting its use as an automated annotation tool in large-scale analyses.

\section{Example}
\label{sec:appendixD}
\begin{center}
\begin{tcolorbox}[breakable,title={Query Example (recheck activation and upper context)}]
{
Then pairwise intersections:

Numbers divisible by 3 and 23: lcm(3,23)=69. floor(2003 / 69). 69*29 = 2001, so floor(2003 / 69) = 29. Correct.

Numbers divisible by 3 and 29: lcm(3,29)=87. floor(2003 / 87). 87*23 = 2001, so floor(2003 / 87) = 23. Correct.

Numbers divisible by 3 and 2003: lcm(3,2003)= 3*2003. Since 3*2003 = 6009 > 2003, floor(2003 / 6009)=0. Correct.

Numbers divisible by 23 and 29: lcm(23,29)=667. floor(2003 / 667). 667*3 = 2001, so floor(2003 / 667)=3. Correct.

Numbers divisible by 23 and 2003: lcm(23,2003)=23*2003, which is way larger than 2003, so 0. Correct.

Numbers divisible by 29 and 2003: same, 0. Correct.

So sum of pairwise intersections: 29 + 23 + 0 + 3 + 0 + 0 = 55. Correct.

Then subtract 55 from 824: 824 - 55 = 769. Correct.

Then triple intersections:

Numbers divisible by 3,23,29: lcm(3,23,29)=3*23*29=2001. floor(2003 / 2001)=1. Correct.

Other triplets: 3,23,2003: lcm=3*23*2003, which is way bigger than 2003, so 0. Similarly others.

So total triple intersections: 1. So add 1: 769 + 1 = 770. Correct.

Four-way intersection: 0. So total numbers not coprime: 770. Therefore, numbers coprime: 2003 - 770 = 1233. 

Wait, but let me check if that answer makes sense.
}
\label{appendix:example_1}
\end{tcolorbox}
\end{center}

\begin{center}
\begin{tcolorbox}[breakable,title={Top-1 Example}]
{

``Context'':\{

So, subtract 55 from the previous total: 824 - 55 = 769.

Now, add back the triple intersections.

Numbers divisible by lcm(3,23,29) = 3*23*29 = 2001. floor(2003 / 2001) = 1.

Numbers divisible by lcm(3,23,2003) = 3*23*2003, which is way larger than 2003, so 0.

Similarly, numbers divisible by lcm(3,29,2003) = 3*29*2003, also larger than 2003, so 0.

Numbers divisible by lcm(23,29,2003) = 23*29*2003, also larger than 2003, so 0.

Therefore, the triple intersections sum up to 1 + 0 + 0 + 0 = 1. So we add 1: 769 + 1 = 770.

Now, subtract the four-way intersection: numbers divisible by lcm(3,23,29,2003) = 3*23*29*2003, which is way larger than 2003, so floor(2003 / that number) = 0. Therefore, subtract 0: 770 - 0 = 770.

Therefore, the total number of numbers not coprime to n is 770. Therefore, the number of numbers coprime to n is total numbers (2003) minus 770, which is 2003 - 770 = Wait, but hold on. Let me check my calculations again, because that seems a bit high. Let me verify each step again.

First, total numbers: 2003. Correct.

First part: numbers divisible by 3, 23, 29, or 2003: 667 + 87 + 69 + 1 = 824. Correct.

Then subtract pairwise intersections:

Numbers divisible by 3 \& 23: 29. 3 \& 29: 23. 3 \& 2003: 0. 23 \& 29: 3. 23 \& 2003: 0. 29 \& 2003: 0. Total pairwise: 29 + 23 + 0 + 3 + 0 + 0 = 55. Correct. So 824 - 55 = 769.

Then add back triple intersections:

Numbers divisible by 3, 23, 29: 2001. floor(2003 / 2001) = 1. The other triplets: 3,23,2003: floor(2003 / (3*23*2003)) = 0. Similarly others. So total triple intersections: 1. So 769 + 1 = 770.

\}

``Label'':\{

``Unnecesasry''

\}

}
\label{appendix:example_2}
\end{tcolorbox}
\end{center}

\section{Prompts}
\label{sec:appendixE}
\begin{center}
\begin{tcolorbox}[title={Reflection Identifier Prompt}]
{
\# Reflection Step Classification Task

**Task:**  
You are given a single step from a model's reasoning process.

**Goal:**  
Decide whether this step is a *reflection step*.

---

\#\# Core Rule (Strict)

A step is a reflection **only if it explicitly refers to the model's own reasoning process** and is about *monitoring, evaluating, or revising that reasoning.*

It must clearly do **at least one** of the following:
- Verify, check, audit, or critique earlier reasoning or results
- Propose changing, reconsidering, or comparing approaches or strategies  

This must be about the **reasoning itself**, not the problem.

---

\#\# Required Signal

The step should contain **explicit self-referential or metacognitive language**, such as:
- "I should check..."
- "This approach might be wrong..."
- "Let me reconsider..."
- "So far, my reasoning assumes..."
- "I may have made a mistake..."
- "Another strategy could be..."

If it does **not** monitor or revise the reasoning process, it is **not** a reflection.

---

\#\# Not a Reflection (Always False)

Do **not** classify as reflection if the step is mainly:
- Solving the problem  
- Doing calculations, algebra, or transformations  
- Applying formulas, rules, or definitions  
- Deriving new results or conclusions  
- Explaining why a step works  
- Stating what to do next without evaluating the reasoning

Even if it uses words like "think", "so", or "therefore", it is **not** a reflection unless it critiques or evaluates the reasoning itself.

---

\#\# Output Format (Strict)

Output only:

True  
or  
False  

No explanations. No extra text.
}
\label{appendix:case_1}
\end{tcolorbox}
\end{center}

\begin{center}
\begin{tcolorbox}[breakable,title={Reflection Type Classifier Prompt}]
{
You are an annotation assistant.

You are given:
- an earlier reasoning step
- several later lines of context from the same reasoning trace
- the reasoning step to be classified

Your task is to classify the step.

You must output exactly one label:
a. Successful Correctness Repair Action  
b. Unsuccessful Correctness Repair Action  
c. Strategy Exploration  
d. Unable to classify  

Only output the single letter.

---

\#\# Core Concepts

\#\#\# A. Correctness Repair Action (CRA)

A step is a Correctness Repair Action IF AND ONLY IF:

1) There exists a prior calculation, result, assumption, bound, interpretation, or conclusion earlier in the trace.

AND

2) The PURPOSE of the step is to evaluate whether that earlier calculation is correct or incorrect.

AND

3) The step performs or initiates a verification action, such as:
- recomputation
- checking constraints
- plugging values back in
- testing cases
- searching for counterexamples
- validating optimality/minimality/maximality

The action must be aimed at CONFIRMING or REFUTING a prior calculation/result.

---

\#\# B. Strategy Exploration

A step is Strategy Exploration IF AND ONLY IF:

1) Its purpose is to propose, compare, or shift solution approaches.

AND

2) It is NOT primarily verifying or refuting a prior belief.

Examples:
- proposing a new method
- suggesting an alternative representation
- reframing how to approach the problem

---

\#\# C. Successful vs Unsuccessful CRA (STRICT)

\#\#\# ABSOLUTE RULE

A step MUST NOT be labeled **Successful Correctness Repair Action** if the resulting calculation/result is the SAME as before.

If the check confirms the earlier calculation/result, it is ALWAYS **Unsuccessful**.

---

\#\#\# Successful Correctness Repair Action (a)

Label **a** if and only if ALL are true:

1) A prior calculation/result in the trace is objectively WRONG.  
2) This step performs a genuine verification or repair attempt.  
3) This step REVEALS the calculation/result is wrong.  
4) The calculation/result is REPLACED with a correct one.  
5) No new error is introduced.

If any condition fails → not successful.

Typical outcomes:
- wrong value recomputed correctly  
- false assumption invalidated  
- flawed argument repaired

---

\#\#\# Unsuccessful Correctness Repair Action (b)

Label **b** if ANY are true:

- The prior calculation/result was actually correct.  
- The step checks something but the conclusion does NOT change.  
- The step fails to detect a real error.  
- The step checks the wrong thing.  
- The step introduces a new error.

This explicitly includes:
- rechecking and getting the same answer  
- confirming a correct result  
- reaffirming a wrong result

Examples (b):
- "Let me recompute... I still get 70."  
- "Plugging it back in, it works."  
- "Double-checking confirms the previous conclusion."

---

\#\# Lexical Guidance (non-binding)

Verification signals:
check, verify, double-check, recompute, re-calculate, plug in, test whether, confirm, see if this holds, counterexample

Strategy signals:
maybe, try, consider, alternatively, another approach, instead

Lexical cues do NOT override outcome rules.

---

\#\# Final decision checklist (must follow):

1) Is this step trying to verify or repair a prior belief?\newline
- No \textrightarrow c or d

2) Is it proposing or comparing approaches rather than checking correctness?\newline
- Yes \textrightarrow c

3) Was there an actually wrong prior belief?\newline
- No \textrightarrow b

4) Did this step expose the error and replace it with a correct belief?\newline
- Yes \textrightarrow a  \newline
- No \textrightarrow b

---

Only output: a, b, c, or d.}
\label{appendix:case_2}
\end{tcolorbox}
\end{center}

\begin{center}
\begin{tcolorbox}[breakable,title={Recheck Activation Extraction Prompt}]
{
You are an annotation assistant. Your task is to extract ONLY sentences where the model checks, verifies, or attempts to falsify the correctness of something it has ALREADY stated.

We annotate correctness repair actions.
We do NOT annotate strategy exploration or forward reasoning.

---

Definition: Correctness Repair Action

A sentence qualifies IF AND ONLY IF:

1) There exists a prior belief, result, assumption, bound, interpretation, or conclusion stated earlier in the rollout (explicitly or implicitly).

2) The sentence’s PURPOSE is to evaluate whether that earlier belief is CORRECT or INCORRECT.

3) The sentence performs or initiates a verification action, which may include:
   - re-computation
   - checking constraints
   - testing candidates
   - searching for counterexamples
   - validating optimality / minimality / maximality
   as long as these actions are used to CONFIRM or REFUTE a prior claim.

---

Lexical guidance (strong signals):

Typical verification phrases:
check, verify, double-check, re-check, confirm, re-calculate, recompute,
plug in, test whether, see if this holds.

If a sentence contains “maybe” or “let me” but does NOT explicitly reference checking correctness (e.g., check / verify / confirm / see if this holds), do NOT annotate it. And do NOT annotate conclusion sentences before starting to check.

---

Examples (annotate these):

1. Re-computation (explicitly redoing a prior step)
- "Wait, let me compute this again carefully."
- "Let me double-check the arithmetic here."

2. Checking constraints (validating a prior result against conditions)
- "Let me check if this satisfies the constraint that X and Y must be inside the square."
- "Let me plug this back into the equation to see if it works."

3. Testing candidates (verifying a prior claim via examples or nearby cases)
- "Let me test a few cases around this value to confirm it’s maximal."
- "Let me try a couple of examples to see if this holds."

4. Searching for counterexamples (attempting to falsify a prior claim)
- "Let me see if I can find a counterexample."
- "Let me check if this ever breaks down."

5. Validating optimality / minimality / maximality
- "Let me check if this is indeed the minimum."
- "Let me verify that no other placement yields a smaller value."

---

Do NOT annotate:

- Summaries, conclusions:
  “Therefore, the answer should be…”, "Therefore, this suggests…"
- Strategy shifts or planning:
  “To solve the prblem, maybe try…”, “Alternatively, consider…”
- Idea generation without correctness intent:
  “Maybe there is a larger number…”
- Forward computation stated for the first time:
  “Compute…”, “We obtain…”, “Thus…”
- Meta-cognition without verification:
  “Let me think again.”

---

Extraction rules:

- Copy the sentence EXACTLY as it appears in the rollout.
- Do NOT annotate fragments or partial clauses.
- Do NOT paraphrase, invent, or duplicate sentences.
}
\label{appendix:case_3}
\end{tcolorbox}
\end{center}

\begin{center}
\begin{tcolorbox}[breakable,title={Recheck Activation Extraction Prompt}]
{
You are an annotation assistant. Your task is to extract ONLY sentences where the model checks, verifies, or attempts to falsify the correctness of something it has ALREADY stated.

We annotate correctness repair actions.
We do NOT annotate strategy exploration or forward reasoning.

---

Definition: Correctness Repair Action

A sentence qualifies IF AND ONLY IF:

1) There exists a prior belief, result, assumption, bound, interpretation, or conclusion stated earlier in the rollout (explicitly or implicitly).

2) The sentence’s PURPOSE is to evaluate whether that earlier belief is CORRECT or INCORRECT.

3) The sentence performs or initiates a verification action, which may include:
   - re-computation
   - checking constraints
   - testing candidates
   - searching for counterexamples
   - validating optimality / minimality / maximality
   as long as these actions are used to CONFIRM or REFUTE a prior claim.

---

Lexical guidance (strong signals):

Typical verification phrases:
check, verify, double-check, re-check, confirm, re-calculate, recompute,
plug in, test whether, see if this holds.

If a sentence contains “maybe” or “let me” but does NOT explicitly reference checking correctness (e.g., check / verify / confirm / see if this holds), do NOT annotate it. And do NOT annotate conclusion sentences before starting to check.

---

Examples (annotate these):

1. Re-computation (explicitly redoing a prior step)
- "Wait, let me compute this again carefully."
- "Let me double-check the arithmetic here."

2. Checking constraints (validating a prior result against conditions)
- "Let me check if this satisfies the constraint that X and Y must be inside the square."
- "Let me plug this back into the equation to see if it works."

3. Testing candidates (verifying a prior claim via examples or nearby cases)
- "Let me test a few cases around this value to confirm it’s maximal."
- "Let me try a couple of examples to see if this holds."

4. Searching for counterexamples (attempting to falsify a prior claim)
- "Let me see if I can find a counterexample."
- "Let me check if this ever breaks down."

5. Validating optimality / minimality / maximality
- "Let me check if this is indeed the minimum."
- "Let me verify that no other placement yields a smaller value."

---

Do NOT annotate:

- Summaries, conclusions:
  “Therefore, the answer should be…”, "Therefore, this suggests…"
- Strategy shifts or planning:
  “To solve the prblem, maybe try…”, “Alternatively, consider…”
- Idea generation without correctness intent:
  “Maybe there is a larger number…”
- Forward computation stated for the first time:
  “Compute…”, “We obtain…”, “Thus…”
- Meta-cognition without verification:
  “Let me think again.”

---

Extraction rules:

- Copy the sentence EXACTLY as it appears in the rollout.
- Do NOT annotate fragments or partial clauses.
- Do NOT paraphrase, invent, or duplicate sentences.
}
\label{appendix:case_4}
\end{tcolorbox}
\end{center}

\begin{center}
\begin{tcolorbox}[breakable,title={Recheck Filtering Prompt}]
{
You are a strict binary classifier.

Task: Given ONE sentence from an LLM rollout, decide whether this sentence is a "correctness repair action onset":
It explicitly initiates or performs a CHECK / VERIFICATION / FALSIFICATION of correctness (including re-computation,
checking constraints, testing candidates to confirm/refute, searching for counterexamples, validating optimality).

Label rules:

- Output "1" only if the sentence itself clearly expresses verification intent (e.g., "check", "verify", "double-check",
  "recompute", "plug in", "test whether this holds", "see if this works", "counterexample", "confirm it's minimal/maximal").
- Output "0" otherwise.

Important:

- Do NOT output 1 for pure planning/strategy ("Maybe try...", "Consider symmetry...", "Let me try coordinates...").
- Do NOT output 1 for generic meta-cognition without checking ("Let me think again.").
- Do NOT output 1 for summaries/conclusions ("Therefore...", "Thus...") unless it explicitly says it is checking/validating.
- Return ONLY a single character: 1 or 0.
}
\label{appendix:case_5}
\end{tcolorbox}
\end{center}

\begin{center}
\begin{tcolorbox}[breakable,title={Outcome Annotation Prompt}]
{
You are an annotation assistant. You will be given:
(1) a recheck onset sentence previously extracted from a rollout, and
(2) its surrounding context (a short window of text before and after).

Your task: Decide whether the recheck that begins at the onset sentence is NECESSARY or UNNECESSARY, using ONLY evidence in the provided context window.

Definitions (outcome-based)

- NECESSARY (corrective): The recheck discovers an error/inconsistency and leads to a correction. Evidence includes: explicit admission of mistake, retraction, changed equation/value, revised claim, changed plan/approach due to the check, or a different conclusion/answer.

- UNNECESSARY (confirmatory): The recheck does not change anything substantive.
  It only confirms correctness, repeats the same derivation/result, or proceeds with the same claim after checking.

- INCONCLUSIVE: The excerpt does not contain enough information to tell whether the check caused a correction or merely confirmed correctness (e.g., the check starts but the outcome is not shown).

Rules:

- Use ONLY the provided context window. Do NOT assume anything outside it.
- Base the decision on explicit textual evidence (e.g., “I was wrong…”, “should be…”, changed numbers,
  “checks out…”, “consistent…”).
- If there is ANY correction caused by the check (even a small arithmetic fix), label NECESSARY.
- If the check finishes and the claim stays the same, label UNNECESSARY.
- If unclear, label INCONCLUSIVE.

}
\label{appendix:case_6}
\end{tcolorbox}
\end{center}




\end{document}